\documentclass[review, a4paper, 11pt]{elsarticle}

\let\today\relax

\usepackage{hyperref}
\usepackage{setspace}
\usepackage[title]{appendix}
\usepackage[margin=3cm]{geometry}

\usepackage{csquotes}
\usepackage{tabularx}
\usepackage{pdflscape}

\usepackage{graphicx}
\usepackage{caption}
\usepackage{subcaption}
\usepackage{color}
\usepackage{hyperref}
\usepackage{mwe}
\usepackage{comment}
\usepackage{amsmath}
\usepackage{amssymb}
\usepackage{amsthm}

\newtheorem{definition}{Definition}
\DeclareMathOperator{\diag}{diag}

\DeclareMathOperator*{\argmax}{arg\,max}
\DeclareMathOperator*{\argmin}{arg\,min}

\usepackage{bm}

\usepackage{enumitem}
\usepackage{mathtools}
\usepackage{algpseudocode}
\usepackage{algorithm}
\usepackage{algorithmicx}
\algnewcommand{\algorithmicxand}{\textbf{ and }}
\algnewcommand{\algorithmicxor}{\textbf{ or }}
\algnewcommand{\OR}{\algorithmicxor}
\algnewcommand{\AND}{\algorithmicxand}
\algnewcommand{\algorithmicxfor}{\textbf{ for }}
\algnewcommand{\algorithmicxdo}{\textbf{ do }}
\algnewcommand{\FOR}{\algorithmicxfor}
\algnewcommand{\DO}{\algorithmicxdo}
\algnewcommand{\algorithmicxbreak}{\textbf{break}}
\algnewcommand{\BREAK}{\algorithmicxbreak}
\newcommand{\algrule}[1][.2pt]{\par\vskip.5\baselineskip\hrule height #1\par\vskip.5\baselineskip}

\usepackage[utf8]{inputenc}
\usepackage{multirow}
\usepackage{dirtytalk}

\usepackage{soul}
\usepackage{xcolor}
\usepackage{verbatim}
\usepackage{multicol}

\def\mbf{\ensuremath\mathbf}

\biboptions{authoryear}

\begin{document}

\begin{frontmatter}

\title{Bi-objective Ranking and Selection Using Stochastic Kriging}


\author[firstaddress,thirdaddress]{Sebastian Rojas Gonzalez\corref{correspondingauthor}}
\cortext[correspondingauthor]{Corresponding author}
\ead{sebastian.rojasgonzalez@ugent.be}

\author[secondaddress]{Juergen Branke}
\author[firstaddress,thirdaddress]{Inneke Van Nieuwenhuyse}

\address[firstaddress]{Department of Information Technology, Ghent University, Belgium.}
\address[secondaddress]{Department of Operations, Warwick University, United Kingdom}
\address[thirdaddress]{FlandersMake@UHasselt and Data Science Institute, Hasselt University, Belgium}

\begin{abstract}

We consider bi-objective ranking and selection problems, where the goal is to correctly identify the Pareto optimal solutions among a finite set of candidates for which the two objective outcomes have been observed with uncertainty (e.g., after running a multiobjective stochastic simulation optimization procedure). When identifying these solutions, the noise perturbing the observed performance may lead to two types of errors: solutions that are truly Pareto-optimal can be wrongly considered dominated, and solutions that are truly dominated can be wrongly considered Pareto-optimal. We propose a novel Bayesian bi-objective ranking and selection method that sequentially allocates extra samples to competitive solutions, in view of reducing the misclassification errors when identifying the solutions with the best expected performance. The approach uses stochastic kriging to build reliable predictive distributions of the objective outcomes, and exploits this information to decide how to resample. Experimental results show that the proposed method outperforms the standard allocation method, as well as a well-known the state-of-the-art algorithm. Moreover, we show that the other competing algorithms also benefit from the use of stochastic kriging information; yet, the proposed method remains superior.  

\end{abstract}


\end{frontmatter}

\section{Introduction}
\label{sec:intro}

\begin{figure}[H]
	\centering
	\includegraphics[width=\textwidth]{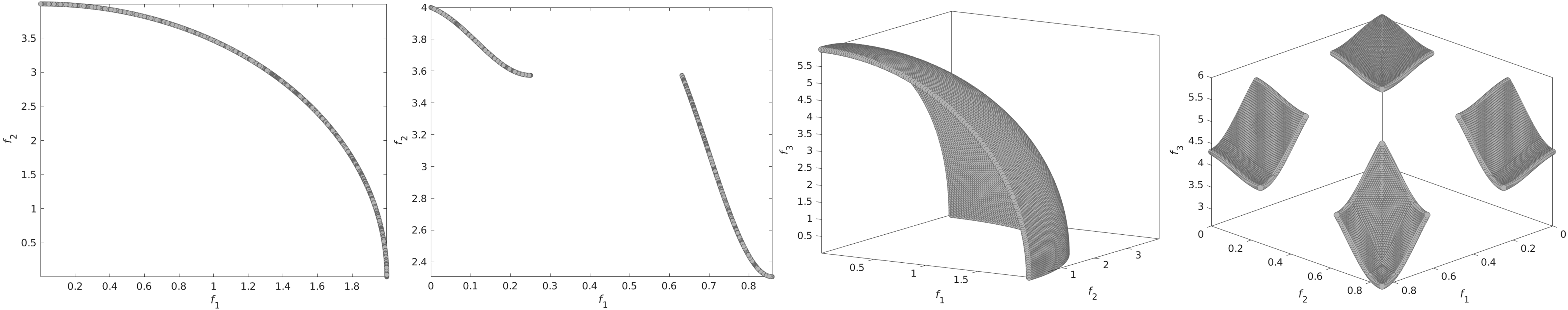}
	\caption{Bi-objective (left) and tri-objective (right) Pareto fronts with different geometries.}
	\label{fig:fronts}
\end{figure}

In \emph{multiobjective} or \emph{multi-criteria} optimization problems, the goal is to find the set of solutions that reveal the essential trade-offs between the objectives (i.e., where no single objective can be improved without negatively affecting any other objective). These solutions are referred to as \emph{non-dominated or Pareto-optimal}, and form the \emph{Pareto set}, also referred to as the \emph{efficient set}; the evaluation of these solutions in the objectives corresponds to the \emph{Pareto front} (see Figure \ref{fig:fronts} for some examples of continuous fronts). Depending on the type of problem, the Pareto front may have different geometries (e.g. concave, convex, linear, disconnected). In real-life problems, the location and geometry of the Pareto front are evidently unknown, and a discrete set of solutions is often used to approximate it \citep{Miet99,Hunter17}.

This article considers problems where the outcomes of the objective functions could only be measured through stochastic experiments (e.g., stochastic simulation), i.e., the observations are noisy. Such problems occur frequently in real life. For instance, in manufacturing and distribution problems (e.g., lot sizing in manufacturing plants), it is common to maximize the expected profit of the production system, while also minimizing profit risk; in transportation problems (e.g., constructing optimal routes for dial-a-ride services) it is common to minimize tardiness of deliveries while also using minimal resources; and in the healthcare sector it is often equally important to minimize the waiting time of patients and the idle time of doctors. We assume that a fixed  set of candidate solutions is available, for which the objective outcomes have been evaluated (with observational noise). Such a set may result, for instance, from running a  multi-objective metaheuristic (see, e.g., \citealt{fieldsend2015rolling}), or a multiobjective simulation optimization procedure (see e.g., \citealt{Li15mo}). The goal of \emph{multi-objective ranking and selection} (MORS) then is to correctly identify the true Pareto-optimal solutions among such a finite set, by adding extra replications to solutions in the most informative way.  

In the current literature on multi-objective stochastic simulation optimization (see e.g., \citealt{horn2017first,Feliot17,pesmo}), this identification phase is often neglected, and the observed mean performance is directly used to determine the Pareto-optimal set. Evidently, sampling variability may then lead to two types of identification errors: designs that are truly Pareto-optimal can be wrongly considered dominated, or designs that are truly dominated can be considered Pareto-optimal. \cite{chen2011stochastic} refer to these errors as \emph{Error Type 1} and \emph{Error Type 2}  respectively, whereas \cite{Hunter17} refer to them as \emph{misclassification by exclusion} (MCE) and \emph{misclassification by inclusion} (MCI). Throughout this paper, we use the MCE/MCI terminology. The accuracy of the objective estimates can obviously be improved by allocating extra samples, yet it is natural to assume that the resampling budget is limited. Allocating this computational budget among the candidates is not trivial. \emph{Static resampling} (allocating the resampling budget uniformly across all alternatives) tends to be inefficient, in particular in settings with limited budget and complex noise structures, as it often wastes a lot of replications on inferior solutions \citep{Seb19}.

Literature on the MORS problem is relatively recent, and still very limited \citep{Hunter17}. In this paper, we propose a sequential MORS method that uses \emph{stochastic kriging} (SK) metamodels \citep{Anke10} to build predictive distributions of the objectives, based on the sample means and variances after replication. This allows the proposed method to exploit correlations in the objective outcomes across different solutions, even in the presence of heteroscedastic noise. Under a few mild conditions, SK metamodels provide reliable predictions that are superior to using sampling information directly \citep{staum2009better}, as the noise in the observed performance is considered in the predictions. We exploit the SK information in a novel heuristic to determine which solutions to resample, and show how other MORS algorithms can also benefit from using such metamodel information. In addition, we propose two screening procedures to reduce the computational burden at each iteration. While some preliminary ideas of the method presented here were published in \cite{Seb19mors}, all the allocation decisions we put forward in this paper are novel and shown competitive against the state-of-the-art. 

The remainder of this article is organized as follows: Section \ref{sec:lr} gives an overview of the relevant literature. Section \ref{sec:probdef} defines the problem and Section \ref{sec:SK} gives the necessary stochastic kriging background. Section \ref{sec:skmors} explains the proposed sampling criteria and screening procedures; the proposed algorithm, referred to as SK-MORS, is outlined in Section \ref{sec:algo}. We design the experiments to evaluate the performance of the algorithm in Section \ref{sec:des_experiments}, analyze the results in Section \ref{sec:experiments} and conclude in Section \ref{sec:conclusion}. 

\section{Related work} \label{sec:lr}

The ranking and selection problem has been widely studied in the single-objective case, and has been approached in different ways. One of the most common goals is to maximize the \emph{probability of correct selection} (PCS), where under some mild conditions, convergence to the solution with the true best expected performance can be theoretically guaranteed (see e.g., \citealt{Boesel03,Kim06,frazier2014fully}). In some cases, however, an extremely large replication budget might be required to differentiate two solutions with almost identical performance, while such small differences in performance are likely not relevant to the decision maker. Therefore, \emph{indifference zone} (IZ) approaches have been developed which give a probabilistic guarantee on the selection of the best solution \emph{or} a solution within a given user-defined quantity from the best (known as the \emph{probability of good selection}). This user-specified quantity defines the indifference zone, and  represents the smallest difference worth detecting (see e.g., \citealt{kim2001fully,Boesel03,fan2016indifference}). 

Relative to the single-objective case, the literature on multiobjective ranking and selection is scarce; a good overview can be found in \cite{Hunter17}. 
The multi-objective PCS (hereafter denoted mPCS) is defined as the probability of correctly identifying the entire Pareto set, and only this set \citep{Branke19}. For both the single and multiobjective case, the true PCS must be estimated, usually via Monte Carlo simulation, which often leads to a computational bottleneck \citep{chen2011stochastic}. The multiobjective case is evidently more complex, as not only the true Pareto front is often continuous and unknown (and thus approximated with a relatively large discrete set), but the dominance relationships between solutions are extremely sensitive to simulation replications (see e.g. the results in \citealt{Seb19}). 

One of the most widely used MORS methods is the \emph{Multiobjective Optimal Computing Budget Allocation} (MOCBA), proposed in \cite{MOCBA}, built upon the well-known single-objective ranking and selection approach OCBA \citep{chen2011stochastic}. As discussed by the authors, the MORS problem can be formulated in several ways (e.g., maximizing the  mPCS or minimizing the  classification errors when identifying the non-dominated set). The proposed allocation rules are complex, and use numerical approximations to estimate which solutions are likely to be dominated. In \cite{chen2011stochastic} a simplified version of the  original MOCBA is proposed, which avoids most of these issues. Another relevant framework is the \emph{SCORE allocations for bi-objective ranking and selection} \citep{Feldman15,Feldman18}, which aims to allocate replications to maximize the rate of decay of the probability of wrongly identifying the true Pareto set, and is asymptotically optimal. Later in \cite{Hunter20}, the work was extended to handle 3 and 4 objectives, showing promising results despite the high computational cost. 

More recently, \cite{androMORS} proposed a multi-objective IZ procedure to estimate the Pareto set with the true best expected performance with statistical guarantees. The work focuses on providing a stopping criterion for replicating designs until a predetermined desired mPCS is achieved, which is desirable in some MORS settings, but at the expense of a large replication budget due to the approximation of several parameters. Another related work appears in \cite{Binois15}, where by means of kriging metamodels, predictive distributions are built over the objectives and conditional simulations are used to estimate the probability that any given point in the objective space is dominated; the method is found to be very sensitive to the (unknown) geometry of the Pareto front, and is computationally expensive. To the best of our knowledge, there are only two other multi-objective IZ procedures in the literature: \cite{mocbaIZ} and \cite{Branke19}. The latter work clearly highlights the shortcomings of the first, but remains limited to the biobjective case, and is computationally demanding. 

A different approach appears in the M-MOBA and M-MOBA-HV algorithms \citep{Branke16,Branke19}, which apply a Bayesian approach to determine the solution that, when replicated further, is expected to yield the maximum \emph{information value} \citep{chick10}. In the M-MOBA algorithm, this solution is the one with the highest probability of changing the current Pareto set, whereas in M-MOBA-HV, it is the one leading to the largest change in the observed hypervolume (a widely used performance metric in deterministic multiobjective optimization, that quantifies the volume of the objective space dominated by a given set of points; see \citealt{Zitz07hyper,auger2012hypervolume} for further details on hypervolume).

Another related stream of research comes from the \emph{evolutionary multiobjective optimization} community, which has to rank the solutions in the population in each iteration, and thus needs to solve a multi-objective ranking and selection problem in each iteration if the underlying optimization problem is noisy. However, this problem is often neglected and the observed mean performance is used to rank the solutions in each iteration. Several authors have looked at formalizing the concept of dominance in noisy multi-objective optimisation (see e.g., \citealt{teich2001pareto,hughes2001evolutionary,fieldsend05,Basseur06}). In \cite{Trautmann09} and \cite{Voss10}, probabilistic dominance is defined by comparing the volume in the objective space of the confidence intervals, and the center point of these volumes is used to determine the dominance relationship. Another related approach is \emph{dynamic resampling}, which varies the additional number of samples based on the estimated variance of the observed objective values. It aims to assess the observed responses at a particular confidence level before determining dominance, and to avoid unnecessary resampling (see e.g., \cite{dsampling,syberfeldt2010evolutionary}). A relatively simple yet well performing approach is to simply allocate additional samples to solutions that survive from one generation to the next, as is proposed in the Rolling Tide Evolutionary Algorithm \citep{fieldsend2015rolling}. 

\section{Problem definition} \label{sec:probdef}

A multiobjective optimization problem can be defined as follows: $\min [f_1(\mathbf{x}),...,f_m(\mathbf{x})]$, for $m = \{2,3\}$ objectives. The solution to this problem is a discrete set of \emph{decision vectors} $\mathbf{x}_i = [x_1,...,x_d]^T, i = 1,...n$, also referred to as \emph{designs} or \emph{points}, which are contained in the decision space $\mathcal{X}$ (usually $\mathcal{X} \subset \mathbb{R}^d$), with $f:\mathcal{X} \rightarrow \mathbb{R}^m$ the vector-valued function with coordinates $f_1,...,f_m$ in the objective space. In the stochastic case, the objectives are perturbed by observational noise: $f_j(\mathbf{x}) = f_j(\mathbf{x}) + \epsilon_j, j = 1,...,m$. The observational noise, denoted by $\epsilon_j, j = 1,...,m$, is commonly assumed to be independent among the different objectives and identically distributed across replications. In practice, the noise tends to be heteroscedastic \citep{Kim06,Anke10}; its level is dependent on the decision variables and thus varies throughout the search space: $f_j(\mathbf{x}) = f_j(\mathbf{x}) + \epsilon_j(\mathbf{x}), j = 1,...,m$. In this paper, we focus on the bi-objective case (i.e., $m = 2$ objectives); to simplify notation,  we sometimes use $f_{ij}$ to denote the noisy performance  $f_j(\mathbf{x}_i) = f_j(\mathbf{x}_i) + \epsilon_j(\mathbf{x}_i)$ of point $\mathbf{x}_i$ on objective $j$. 

In a multi-objective setting, usually the concept of \emph{dominance} is used to compare solutions:

\theoremstyle{definition}
\begin{definition}{For $\mathbf{x}_1$ and $\mathbf{x}_2$ two vectors in $\mathcal{X}$:} \label{def:par_dom}
\begin{itemize}
	\item $\mathbf{x}_1 \prec \mathbf{x}_2$ means $\mathbf{x}_1$ dominates $\mathbf{x}_2$ iff $f_j (\mathbf{x}_1) \leq f_j (\mathbf{x}_2),  \forall j \in \{1,..,m\}$, and $\exists j \in \{1,..,m\}$ such that $f_j (\mathbf{x}_1) < f_j (\mathbf{x}_2) $
	\item $\mathbf{x}_1 \prec \prec \mathbf{x}_2$ means $\mathbf{x}_1$ strictly dominates $\mathbf{x}_2$ iff $f_j (\mathbf{x}_1) < f_j (\mathbf{x}_2), \forall j \in \{1,..,m\}$.
\end{itemize}
\end{definition}

\noindent The set of solutions not dominated by any other solution is called the \emph{Pareto set}, its corresponding image in the objective space is called \emph{Pareto front}. In the noisy case, the observed performance for a given objective at a given decision vector $\mathbf{x}_i$ is commonly estimated by the sample mean of that objective over $r_i$ replications: $\bar{f}_{ij} = \sum_{k=1}^{r_i} \bar{f}_{ij}^k/r_i$, $j = 1,...,m$, where $\bar{f}_{ij}^k$ denotes the performance of the k-th replication of $\mathbf{x}_i$ on objective $j$. These sample means are then directly used to obtain the \emph{observed} Pareto front, which clearly may incur MCE/MCI errors. 

We thus consider the problem of efficiently identifying the designs with the true best expected performance out of a given set of alternatives, when these alternatives are evaluated on multiple objectives using a stochastic simulator. As common in the literature, we assume that the true performance measure $f_{ij}$ for any arbitrary objective $j$ in design $i$ follows a normal prior distribution, and no prior knowledge is available on the performance before conducting the simulation studies.

\begin{figure}[h]
	\centering
	\includegraphics[width=\textwidth]{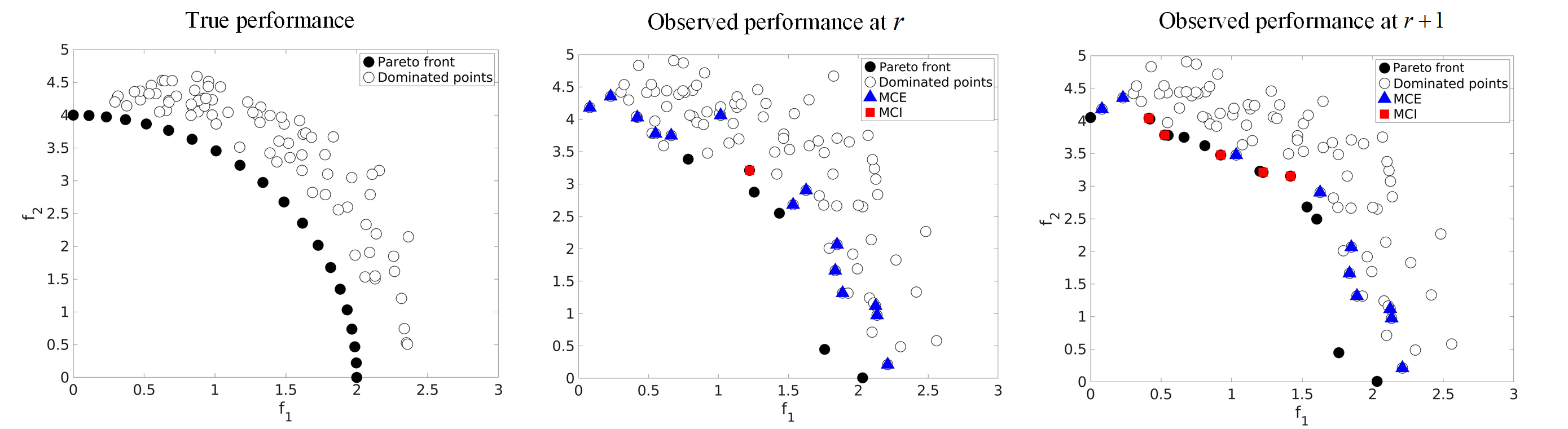}
	\caption{Left panel: True performance of a given set of solutions, for a bi-objective minimization problem. Center panel: Observed performance based on the sample means after $r$ replications per solution. Right panel: Observed performance after $r+1$ replications per solution.}
	\label{fig:reps}
\end{figure}

Given $r_i$ independent and identically distributed replications (iid) of design $i$, the posterior distribution of the performance is then commonly estimated by

\begin{equation}
\{f_{ij}\} 
{\sim} \mathcal{N}(\bar f_{ij},\frac{s^2_{ij}}{r_i}), \forall i = 1,..,n; j = 1,...,m
\label{eq:normal}
\end{equation}

\noindent where $\bar{f}_{ij}$ and $s^2_{ij}$ denote the sample mean and sample variance of objective $j$ in design $i$, respectively. If every possible solution has been evaluated at least once, the \emph{observed} Pareto set (i.e., the Pareto set based on sample means) can be determined  based on the $r_i, i = 1,...,n$ replications available so far on each objective. After an additional $r^{\prime}_i$ simulation replications are performed, yielding an average performance of $\bar{f}^{\prime}_{ij}$, the observed mean is updated as \citep{Branke19}:

\begin{equation}
\bar{f}_{ij} = \frac{r_i\bar{f}_{ij} + r_i^{\prime} \bar{f}^{\prime}_{ij}}{r_i+r^{\prime}_i}. 
\end{equation}
\noindent Naturally, when the observed means are updated, the dominance relationships among some alternatives will most likely change, and in some cases change drastically (see Figure \ref{fig:reps}), thereby also changing the observed Pareto set. We thus focus on solving the following optimization problem (see also \citealt{MOCBA}): 
\begin{align} \label{eq:opt_prob}
&\min_{r_1,...,r_n} (\text{MCE+MCI})          \\   
\text{s.t.}& \sum_{i=1}^n r_i \leq R    \notag \\
&r_i \geq 0, i = 1,...,n                \notag
\end{align}
\noindent where $r_i$ denotes the number of replications on design $i$, $R$ is the total replication budget available, and MCE and MCI are the number of misclassification errors by exclusion and inclusion, respectively. Problem \ref{eq:opt_prob} is analytically intractable: the exact number of errors cannot be computed, as the true Pareto optimal points are in general unknown. Furthermore, the observed sample means are random variates themselves, given that they are estimated from the samples. In this paper we present a sequential heuristic that aims to solve the problem using \emph{stochastic kriging metamodels} \citep{Anke10}. In what follows, we first present the important background of stochastic kriging in Section \ref{sec:SK}, succeeded by the description of the proposed approach in Section \ref{sec:skmors}.  

\section{Stochastic kriging} \label{sec:SK}

\noindent Let $f(\mbf{x})$ be an unknown objective function, which can only be observed with (heteroscedastic) noise:
\begin{align}
\tilde{f}(\mbf{x}_i) = f(\mathbf{x}_i) + \epsilon(\mathbf{x}_i)
\label{fbar}
\end{align}
Assume that we have observed this function in a finite set of $d$-dimensional designs $\mathbf{x}_i, i = 1,...,n$, using $r_i$ replications per design. {\bf Stochastic kriging (SK)} \citep{Anke10} then represents the simulation output in replication $k$ by the following model:

\begin{align}  
\tilde{f}^k(\mbf{x}_i) &= \beta + M(\mbf{x}_i) + \epsilon^k(\mbf{x}_i).\label{sec2:eq:sk}
\end{align}

\noindent where $\beta$ is a constant, and $M(\mbf{x}_i)$ is a realization of a mean 0 random field. The random field $M$ is assumed to exhibit spatial correlation, meaning that the outcomes $M(\mbf{x}_i)$ and $M(\mbf{x}_h)$ will tend to be similar when $\mbf{x}_i$ is close to $\mbf{x}_h$ in the design space. \cite{Anke10} refer to the random nature of $M$ as \emph{extrinsic} uncertainty, as it is imposed on the problem in view of developing the model. In the design and analysis of computer experiments, the standard assumption for $M$ is that it is a Gaussian random field. Note that, in stochastic simulation, the \emph{intrinsic} noise terms in Eq. \ref{sec2:eq:sk} are naturally independent and identically distributed across replications, with $E[\epsilon^k(\mbf{x}_i)]=0$ and $Var[\epsilon^k(\mbf{x}_i)]=\tau^2(\mbf{x}_i)$; as $\tau^2(\mbf{x}_i)$ depends on the location of $\mbf{x}_i$, the model naturally allows for heterogeneous noise. 

The core of the extrinsic spatial correlation model is the \emph{covariance function} or \emph{kernel}, which is used to quantify the covariance between the outcomes of any two decision vectors $\mbf{x}_i$ and $\mbf{x}_h$. In this work we use the (stationary) squared exponential kernel, also known as Gaussian kernel (for further details on other popular kernels, see \cite{Rasm06}, Chapter 5): 
\begin{align}
cov[M(\mbf{x}_i),M(\mbf{x}_h)] = v^2\exp\left[-\sum_{q=1}^{d}\left(\frac{|x_{i,q}-x_{h,q}|}{l_q} \right)^2\right] \label{eq:gauss} 
\end{align}
\noindent where $v^2$ and $l_q, q=1,...,d$ are  \emph{hyperparameters} that denote the process variance and the length-scale of the process along dimension $q$, respectively, for all $i,h = 1,...,n$ and $i \neq h$. These hyperparameters, along with the constant $\beta $ in Eq. \ref{sec2:eq:sk}, need to be estimated from the available data (i.e., the finite set of design points and their corresponding sample means), usually by means of \emph{maximum likelihood estimation} (MLE).

Using the resulting MLE estimates $\hat{\beta}$, $\hat{v}^2$ and $\hat{l}_q, q=1,...,d$, the analyst can predict the outcome for the objective function in Eq. \ref{fbar}, at any arbitrary $\mbf{x}_i$, by means of the \textbf{stochastic kriging predictor}:

\begin{align}  
\hat{f}(\mbf{x}_i) =& \hat{\beta} + \Sigma_M(\mbf{x}_i,\cdot)^T[\Sigma_M+\Sigma_\epsilon]^{-1}(\mbf{\bar{f}}-\hat{\beta}\mbf{1}_n) \label{sec2:eq:sk_pred} 
\end{align}

\noindent where  $\mbf{\bar{f}}= [\bar{f}(\mbf{x}_1),...,\bar{f}(\mbf{x}_n)]^T$ is the $n\times1$ vector containing the sample means of the available solutions, and $\mbf{1}_n$ is a $n\times1$ vector of ones. The $n \times n$ matrix $\Sigma_M$ contains the kernel results for each couple of available points (i.e., $cov[M(\mbf{x}_h),M(\mbf{x}_{h'})]$, for $h,h'=1,...,n$). Analogously, the $n \times 1$ vector $\Sigma_M(\mbf{x}_i,\cdot)$ contains the kernel results for the given point $\mbf{x}_i$ and each of the $n$ already sampled points (i.e.,  $cov[M(\mbf{x}_i),M(\mbf{x}_h)]$, for $h=1,...,n$). The $n \times n$ matrix $\Sigma_{\epsilon}$ contains the covariances implied by the intrinsic noise at the available points. As discussed in \cite{Anke10}, it is in general not recommended to use common random numbers (CRN) in the simulations when stochastic kriging is used (see also \cite{Chen2012} for further details); consequently, $\Sigma_{\epsilon}$ reduces to the diagonal matrix $\diag[\tau^2(\mbf{x}_1)/r_1,...,\tau^2(\mbf{x}_n)/r_n]$. The {\bf predictor uncertainty} is quantified by its mean squared error (\emph{MSE}), and is given by
\begin{align} 
\hat{s}^2(\mbf{x}_i) =& \Sigma_M(\mbf{x}_i,\mbf{x}_i) - \Sigma_M(\mbf{x}_i,\cdot)^T[\Sigma_M+\Sigma_\epsilon]^{-1}\Sigma_M(\mbf{x}_i,\cdot)+\frac{\gamma^T\gamma}{\mbf{1}_n^T[\Sigma_M+\Sigma_\epsilon]^{-1}\mbf{1}_n} \label{sec2:eq:sk_var} \\
\text{with } \gamma =& 1-\mbf{1}_n^T[\Sigma_M+\Sigma_\epsilon]^{-1}\Sigma_M(\mbf{x}_i,\cdot) \notag.
\end{align}

Note that when $\Sigma_\epsilon$ reduces to a zero matrix, the stochastic kriging estimates in Eq. \ref{sec2:eq:sk_pred} and Eq. \ref{sec2:eq:sk_var} reduce to the well-known \emph{ordinary kriging} expressions \citep{jones1998efficient,Kleijnen15}, as the output observations are then essentially deterministic. When that happens, the kriging \emph{predictor} will perfectly coincide with the sample means at all simulated design points, and the \emph{predictor uncertainty} will reduce to zero. Yet, in stochastic simulation, this would require all points in the design to be evaluated with an infinite replication budget. Consequently, in all practical settings, the SK predictor in Eq. \ref{sec2:eq:sk_pred} will \emph{not} coincide with the sample means at the observed points; yet, the difference between them should \emph{tend} to zero as more replications are added. We exploit this crucial property in our proposed ranking and selection procedure, as discussed in the next section.

\section{Proposed MORS procedure} \label{sec:skmors}

\subsection{Proposed sampling criteria} \label{sec:sampling} 

Let $S$ be the entire set of sampled points. Relying solely on sample means, the analyst could detect an ``observed" Pareto set (denoted $PS$) and a corresponding ``observed" Pareto front (denoted $PF$), among the points in $S$. Relying on the SK predictors, also a ``predicted" $\widehat{PS}$ (and corresponding $\widehat{PF}$) could be distinguished. Given the presence of noise, the predicted and observed sets (and fronts) are typically not the same; yet, adding more replications will likely reduce the noise, thus reducing the difference between the sample means and the SK predictors, as argued at the end of the previous section. As opposed to previous work, the sampling criteria proposed in our approach rely on \emph{both} types of information, to make decisions on where to allocate extra replications. The proposed criteria are referred to as \emph{expected hypervolume difference} (EHVD), and \emph{posterior distance} (PD). For ease of notation (analogous to the sample means and variances), for point $\mbf{x}_i$ on objective $j$, hereafter the stochastic kriging predictor and predictor uncertainties are denoted $\hat{f}_{ij}$ and $\hat{s}_{ij}$, respectively. Furthermore, $\mathbf{\bar{f}}_i$ and $\mathbf{\hat{f}}_i$ denote the sample mean objective vectors and predicted objective vectors, respectively. 

\subsubsection{Expected hypervolume difference} \label{subsec:EHVD}

\begin{figure}[h]
	\centering
	\includegraphics[width=0.8\textwidth]{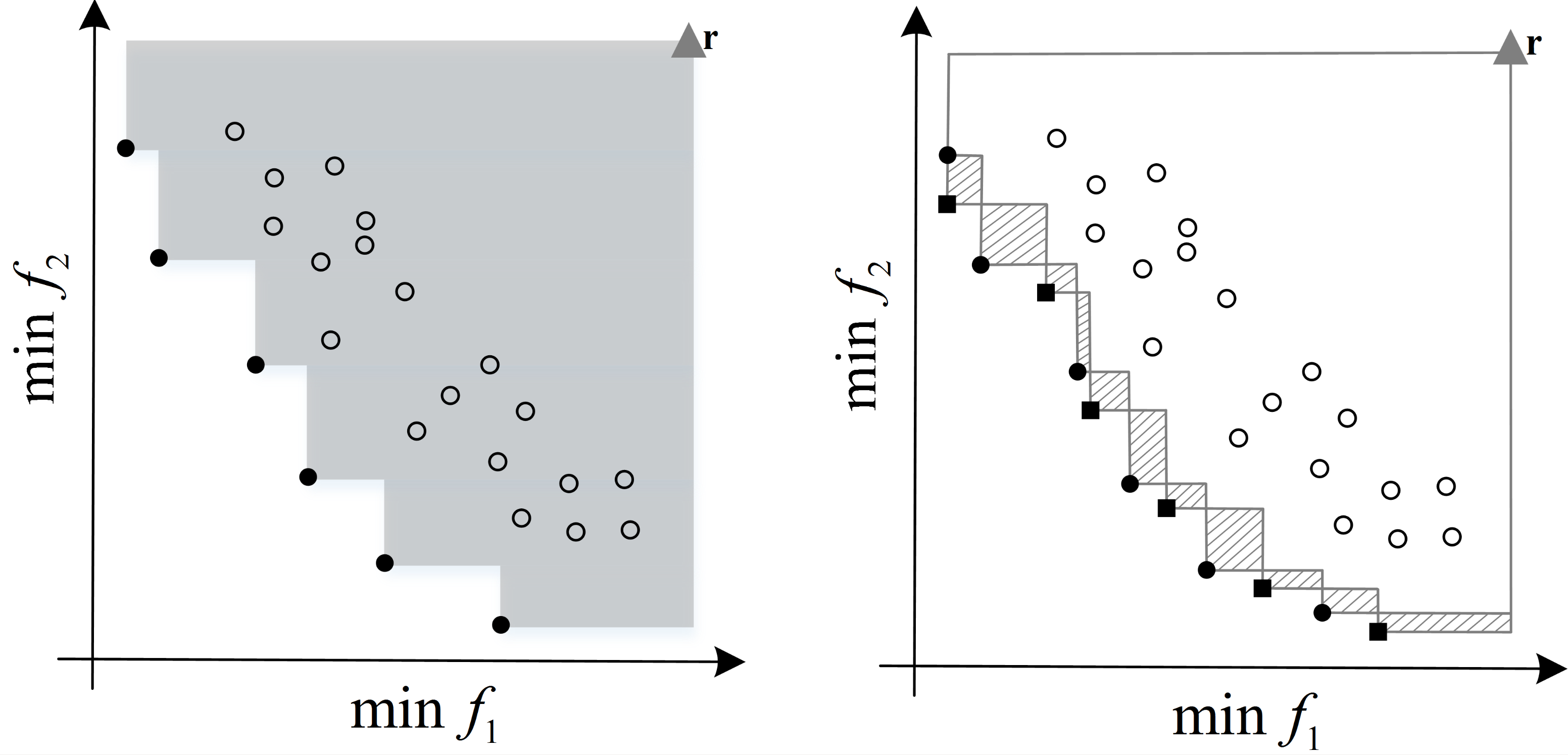}
	\caption[Hypervolume and hypervolume change]{Left panel: Hypervolume (shaded area) dominated by a given non-dominated set (filled points) with respect to a reference point $\mathbf{r}$. Right panel: the EHVC (shaded area) of a set of non-dominated points with respect to a reference point $\mathbf{r}$. The filled circles denote the current observed front (set P), and the filled squares the expected new performance after new samples are allocated (set Q). Empty circles represent the dominated points.}
	\label{fig:hv}
\end{figure}

The hypervolume dominated by a given Pareto front $P$ with respect to a reference point $\mbf{r}$ is defined as the Lebesgue measure, denoted $\Lambda$, of the set of objective vectors dominated by the solutions in $P$, but not by $\mbf{r}$:
\begin{equation}
HV(P,\mbf{r}) = \Lambda \left(\bigcup_{z \in P} \{z': z \prec z' \prec \mbf{r} \} \right). \label{sec2:eq:hv}
\end{equation}
\noindent Thus, all the non-dominated vectors contribute to the indicator value, and the dominated vectors do not contribute. For two fronts P and Q and a reference point $\mbf{r}$, the expected hypervolume change \citep{Branke19} is then defined as:
\begin{equation}
EHVC(P,Q,\mbf{r}) := HV(P,\mbf{r}) + HV(Q,\mbf{r}) - 2 \times \Lambda \left(HV(P,\mbf{r}) \cap HV(Q,\mbf{r})\right). \label{eq:hvd}
\end{equation}
\noindent Figure \ref{fig:hv} illustrates the concept; for a given Pareto front (set P) based on sample means (left panel), the EHVC calculates how much this set is expected to change (set Q), under the assumption that the performance of each solution follows a certain probability distribution.

The EHVC was proposed in \cite{Branke16} and \cite{Branke19} as a criterion to allocate samples. Analytical expressions for EHVC were proposed for the bi-objective case; yet, computational complexity remains high. We use the concept of EHVC to estimate the expected change in the observed HV, if more samples were allocated to a given point $\mbf{x}_i$, by replacing the observed objective outcomes (i.e., $\mathbf{\bar{f}}_i$) by their SK predictors (i.e., $\mathbf{\hat{f}}_i$), as we expect these predictors to yield more accurate estimates of the true objective values. We refer to this criterion as the \emph{expected hypervolume difference} (EHVD):
\begin{equation}
\text{EHVD}_i = \mid HV(PF) - HV\left(PF \backslash \mathbf{\bar{f}}_i \cup \mathbf{\hat{f}}_i\right)\mid, \hspace{1cm} \forall i \in S. \label{eq:ehvc}
\end{equation}

\begin{figure}[h!]
	\centering
	\includegraphics[width=0.85\textwidth]{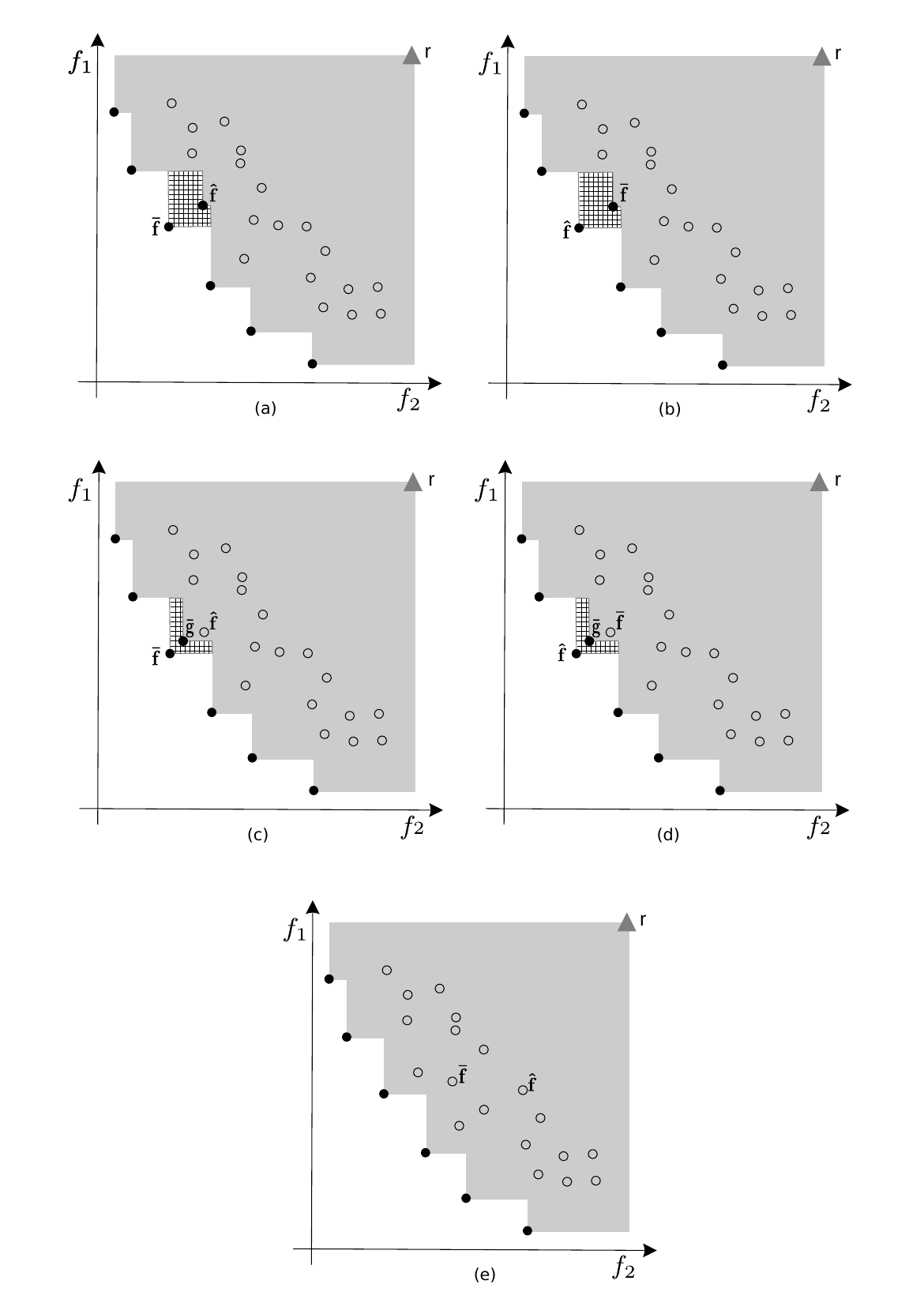}
	\caption{The five cases to be considered when calculating the EHVD.}
	\label{fig:cases}
\end{figure}

\noindent As mentioned before, $\mathbf{\bar{f}}_i$ and $\mathbf{\hat{f}}_i$ will converge as the number of replications for design $i$ grows, and the noise is reduced. As the computation of the HV only uses the non-dominated points, there are five cases to be considered:
\begin{itemize}[noitemsep]
\item Case 1: $\mathbf{\bar{f}}_i \prec \mathbf{\hat{f}}_i$, and $\mathbf{\hat{f}}_i$ is on the Pareto front. The HV will change (decrease) as the observed vector dominates the predicted vector (see Figure \ref{fig:cases}(a)).
\item Case 2: $\mathbf{\hat{f}}_i \prec \mathbf{\bar{f}}_i$, and $\mathbf{\hat{f}}_i$ is on the Pareto front. The HV will change (increase) as the predicted vector dominates the observed vector (see Figure \ref{fig:cases}(b)).
\item Case 3: $\mathbf{\bar{f}}_i \prec \mathbf{\hat{f}}_i$, and there are one or more points that dominate $\mathbf{\hat{f}}_i$ (denoted $\mbf{\bar{g}}$ in Figure \ref{fig:cases}(c)). The HV will change (decrease) as the sample mean vector dominates the predicted vector, but the change is not equally significant as in Case 1.
\item Case 4: $\mathbf{\hat{f}}_i \prec \mathbf{\bar{f}}_i$, and there are one or more points that dominate $\mathbf{\bar{f}}_i$ (denoted $\mbf{\bar{g}}$ in Figure \ref{fig:cases}(d)). The HV will change (increase) as the predicted vector dominates the sample vector, but the change is not equally significant as in Case 2.
\item Case 5: $\mathbf{\bar{f}}_i \prec \mathbf{\hat{f}}_i$ or viceversa, and both performance vectors are observed in the dominated space. The HV will thus not change and will not impact the EHVD (see Figure \ref{fig:cases}(e)).
\end{itemize}

Clearly, the EHVD will tend to allocate more budget to points belonging to Cases 1 and 2; points belonging to Cases 3-4 may have little or no replications allocated, while points in Case 5 will not be considered, though they may as well belong to the true non-dominated set. This is true especially when the noise is high and/or strongly heterogeneous, and thus the sample and predicted means are both relatively far from the true performance; indeed, an extra replication on a given point might then significantly change its position in the objective space.

\subsubsection{Posterior distance} \label{subsec:PD}

The EHVD looks only at the points that are likely to change the current HV the most. In a ranking and selection context, all the points sampled have a probability of belonging to the true Pareto front, as they are relatively close to each other. Thus, we propose a criterion to improve the position in objective space of \emph{any} point in $S$. We define the \emph{posterior distance (PD)} between the observed and predicted means as:
\begin{equation}
\text{PD}_i = \sqrt{\sum_{j=1}^m\left[\mid \bar{f_j}(\mathbf{x}_i)-\hat{f_j}(\mathbf{x}_i) \mid + \hat{s}_{ij} \right]^2}, \hspace{1cm} \forall i \in S. \label{eq:pd}
\end{equation}

\begin{figure}[h!]
	\centering
	\includegraphics[width=\textwidth]{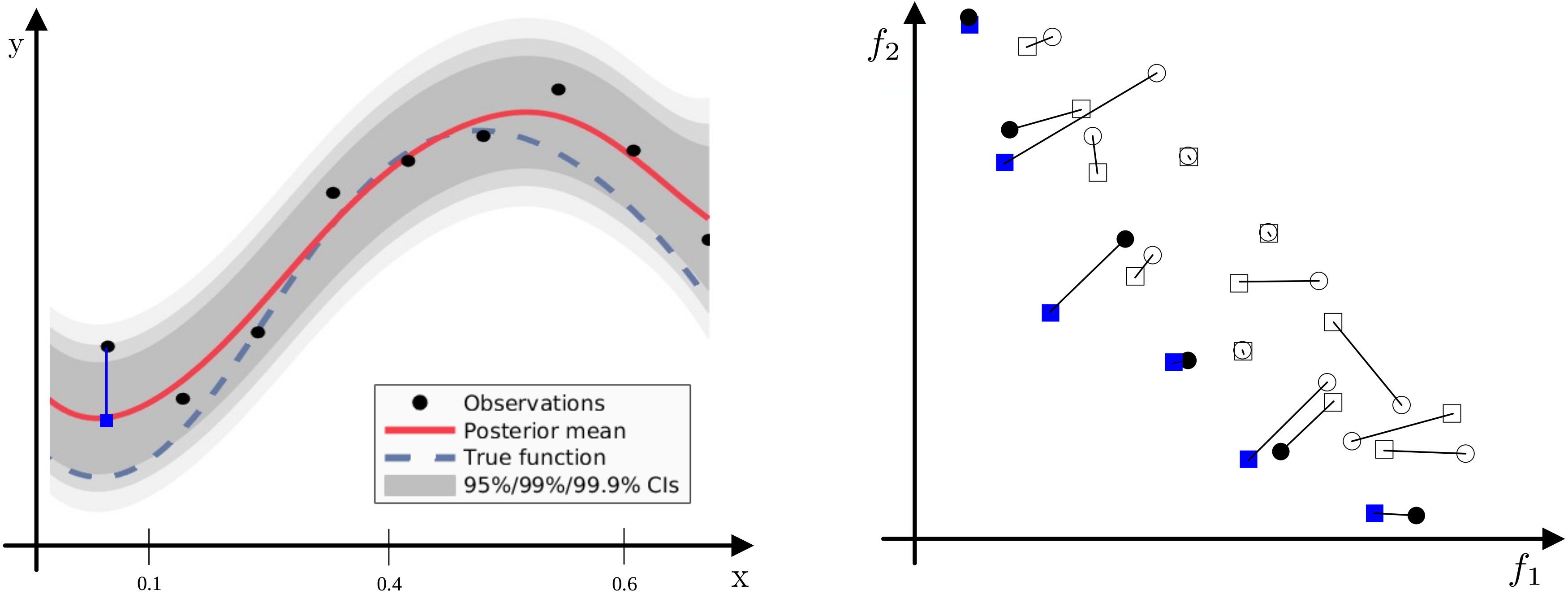}
	\caption{Left panel: Sample means (black dots), posterior means (red line) and posterior uncertainty (gray area) for a single objective. The distance between the prediction and the sample mean is denoted with a blue line. Right panel: The posterior distance between the sample means (circles) and predicted means (squares). The observed and predicted fronts are depicted with filled circles and squares respectively.}
	\label{fig:ppd}
\end{figure}

The PD is the Euclidean distance between the sample means and the predicted means (also referred to as the \emph{posterior means} of the model). If the sample mean and prediction are far away from each other, by running more replications on these points, we expect them to come closer to each other and move towards the true position of the point in the objective space, which aids in minimizing \emph{both} MCE and MCI errors. Note that in Equation \ref{eq:pd}, we add the posterior uncertainty to the distance between the means, as we intend to inflate the PD when the uncertainty of the predictor is high, such that points with high uncertainty will be rewarded more (see Figure \ref{fig:ppd} for an illustration). 

\subsection{Proposed screening procedures} \label{screening}

As is well-known in ranking and selection, in many cases some of the points included in the solution set will be clearly inferior to other solutions in the set \citep{Boesel03}. \emph{Screening} or \emph{subset-selection} heuristics are widely used prior to running the ranking and selection procedure in order to filter out these clearly inferior solutions. We propose two subset selection procedures that will exploit both the sample variance and prediction uncertainties to enclose confidence regions that reduce the number of points to be considered at each iteration. To do this, we use the \emph{lower confidence bounds (LCB)} and \emph{upper confidence bounds (UCB)} of the mean performance, defined as $\bar{f}_j(\mathbf{x}) \pm \omega s_j(\mbf{x}), j = 1,...,m$, where $\omega$ is usually a number in the interval $[1,3]$, to yield the 60\%-99\% confidence intervals (CI) \citep{Rasm06}. Analogously, the CI of the predicted means (denoted $\widehat{LCB}$ and $\widehat{UCB}$) are defined as $\hat{f}_j(\mathbf{x}) \pm \omega\hat{s}_j(\mbf{x}), j = 1,...,m$. Unless stated otherwise, hereafter we will use a value of $\omega = 3$, in order to enclose the 99\% CI (i.e., 3 standard deviations from the mean). 

\begin{figure}[h!]
	\centering
	\includegraphics[width=0.8\textwidth]{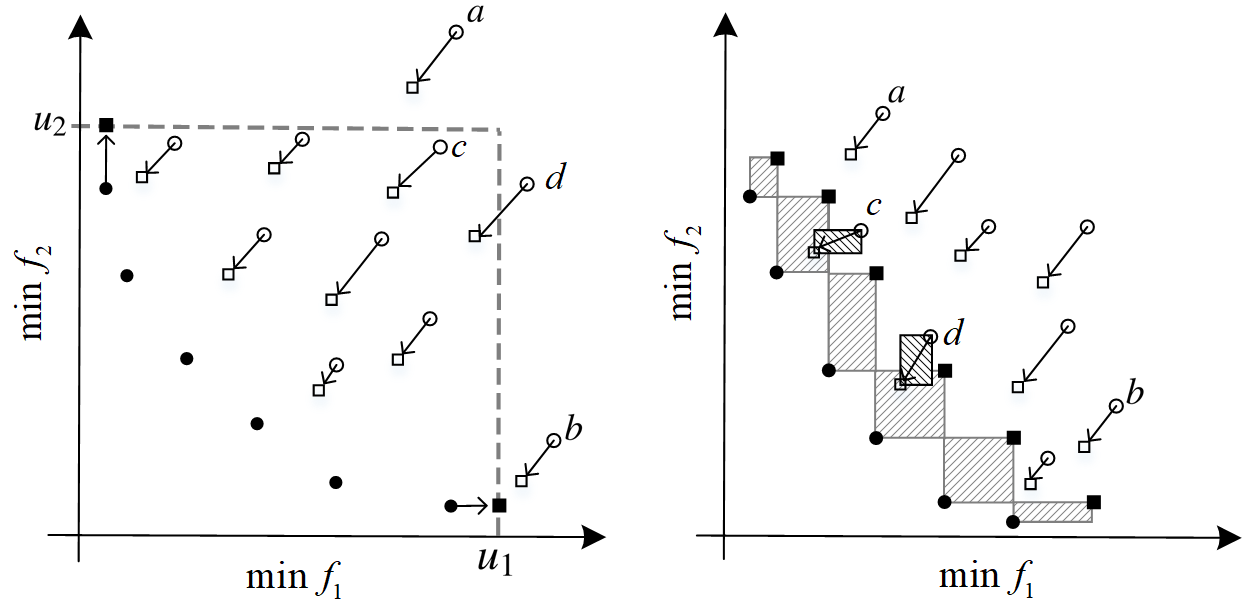}
	\caption[Screening procedures]{Illustration of both screening procedures proposed: \textsc{Screening Box} (left panel) and \textsc{Screening Band} (right panel). Filled marks denote the non-dominated points, and unfilled marks the dominated points. The performance is denoted with circles, and confidence bounds with squares.}
	\label{fig:screening}
\end{figure}

The first of the proposed screening procedures is outlined in Algorithm \ref{alg:screening1}. The worst confidence bounds (i.e., the side of the bound that worsens the performance) among all the non-dominated points for each objective $j$, denoted $\bar{u}_j$ for the sample means, and $\hat{u}_j$ for the predicted means, are used to enclose a confidence region (see left panel of Figure \ref{fig:screening}). The second screening procedure, outlined in Algorithm \ref{alg:screening2}, will instead enclose a confidence region using the upper confidence bounds of \emph{all} the non-dominated points (see right panel of Figure \ref{fig:screening}). These confidence regions are computed for \emph{both} the $PF$ and $\widehat{PF}$. 

\begin{algorithm} [h!]
	\caption{\textsc{Screening Box}}\label{alg:screening1}
	\begin{algorithmic}[1]
		\State \textbf{Input:} $UCB$, $LCB$, $\widehat{UCB}$ and $\widehat{LCB}$
		\State $\bar{S} = \emptyset$ \Comment{Initialize set of clearly inferior points}
		\State $\bar{u}_j$: $\max_{i \in PF} UCB_i, j = 1,...,m$
		\State $\hat{u}_j$: $\max_{i \in \widehat{PF}} \widehat{UCB}_i, j = 1,...m $
		\State $NP = S \setminus PS \cap S \setminus \widehat{PS}$.
 		\For {$i \in NP$}
			\For {$j = 1:m$}
				\If {$LCB_{ij} > \bar{u}_j$ \AND $\widehat{LCB}_{ij} > \hat{u}_j$}	
					\State $\bar{S} \cup \{\mbf{x}_i\}$
					\State \BREAK
				\EndIf
			\EndFor
		\EndFor
		\State \textbf{return} $\bar{S}$
	\end{algorithmic}
\end{algorithm}

\begin{algorithm} [h!]
	\caption{\textsc{Screening Band}}\label{alg:screening2}
	\begin{algorithmic}[1]
		\State \textbf{Input:} $UCB$, $LCB$, $\widehat{UCB}$ and $\widehat{LCB}$
		\State $\bar{S} = \emptyset$ \Comment{Initialize set of clearly inferior points}		
		\State $NP = S \setminus PS \cap S \setminus \widehat{PS}$
		\Procedure{Distance}{$\mathbf{f}_i, \mathbf{f}_k$}
			\State $h_{ik} = \sqrt{\sum_{j=1}^m [f_{ij}-f_{kj}]^2}$
			\State \textbf{return} $h_{ik}$
		\EndProcedure 
		\For {$i \in NP$}
 			\State $\bar{h}_{ik} = $\textsc{Distance}($LCB_i, UCB_k$), $\forall k \in PF$
 			\State $\hat{h}_{ik} = $\textsc{Distance}($\widehat{LCB}_i, \widehat{UCB}_k$), $\forall k \in \widehat{PF}$
		\EndFor
 		\For {$i \in NP$}
 			\State $u = \argmin_{k \in PF} \bar{h}_{ik}$
			\State $v = \argmin_{k \in \widehat{PF}} \hat{h}_{ik}$
			\For {$j = 1:m$}
				\If {$LCB_{ij} > UCB_{uj}$ \AND $\widehat{LCB}_{ij} > \widehat{UCB}_{vj}$}	
					\State $\bar{S} \cup \{\mbf{x}_i\}$
					\State \BREAK
				\EndIf
			\EndFor
		\EndFor
		\State \textbf{return} $\bar{S}$
	\end{algorithmic}
\end{algorithm}

Subsequently, the performance of every dominated mean objective point (visualized with empty circles) is substituted by the lower confidence bounds (visualized with empty squares). Similarly, the performance of the non-dominated points (denoted with filled circles) is substituted by the upper confidence bounds (denoted with filled squares). This is done for both, the performance based on sample means and the performance based on predicted means. If the new position of a given dominated point does not enter any of the confidence regions enclosed by the non-dominated points (i.e., the region enclosed by $PF$ or $\widehat{PF}$), then this point is screened out from the \emph{current} iteration  (see e.g., points $a$ and $b$ in Figure \ref{fig:screening}). Otherwise, it will be considered in the current iteration (see e.g., points $c$ and $d$ in Figure~\ref{fig:screening}). This is to ensure that, even with very high noise levels (as in the experiments in Section \ref{sec:experiments}), with a high confidence, we are not excluding truly non-dominated solutions. 

\subsection{Algorithm outline} \label{sec:algo}

The general steps of the proposed MORS procedure are summarized in Algorithm \ref{alg:algo}; for ease of reference, we summarize the different sets of points used in Table \ref{table:notation}. The initial step of algorithm fits a SK metamodel to each objective based on sample means and variances of the points in $S$, after $b_0$ replications performed in each point. The metamodels are then used to predict each response and respective prediction errors on these sampled points. The current observed Pareto-optimal sets are obtained based on both the sample means ($PS$ and $PF$) and predicted means ($\widehat{PS}$ and $\widehat{PF}$). In the next step, the upper and lower confidence bounds for each point are calculated. Using these confidence bounds, if a screening procedure is invoked (lines 11-13 in Algorithm \ref{alg:algo}), the algorithm filters out from the \emph{current} iteration the points that are observed as clearly inferior. Note that these points might be considered in further iterations.  

\bgroup
\def\arraystretch{0.9}
\begin{table}[h!]
\centering
\caption{ Overview of the different sets of points used in the proposed algorithms.}
\begin{tabular}{ |c|l| }
\hline
Notation & Description\\
\hline
\hline
$S$ & Entire set of sampled points. \\
\hline
$PS$ & Pareto set based on sample means. \\
\hline
$PF$ & Pareto front based on sample means. \\
\hline
$\widehat{PS}$ & Pareto set based on predicted means. \\
\hline
$\widehat{PF}$ & Pareto front based on predicted means. \\
\hline
$UCB$ & Set of the upper confidence bounds of all points in $PS$. \\
\hline
$\widehat{UCB}$ & Set of the upper confidence bounds of all points in $\widehat{PS}$. \\
\hline
$LCB$ & Set of the lower confidence bounds of all points in $S \backslash PS$. \\
\hline
$\widehat{LCB}$ & Set of the lower confidence bounds of all points in $S \backslash \widehat{PS}$. \\
\hline
\end{tabular}
\label{table:notation}
\end{table}
\egroup

\begin{algorithm} [h!]
	\caption{SK-MORS algorithm}\label{alg:algo}
	\begin{algorithmic}[1]
		\State \textbf{Input:}
		\State $S$ \Comment{Set of candidate points}
		\State $b_0$ \Comment{Initial replication budget per point}
            \State $b_{max}$ \Comment{Maximum replication budget per point}
		\State $B$  \Comment{Replication budget to be allocated per iteration}
    	\State \textbf{Output:}
    	\State $PS$ \Comment{The observed Pareto set}
            \algrule
            \State Perform $b_0$ replications on all points in $S$
            \State $S_{max} = \emptyset$ \Comment Initialize set of points that reached $b_{max}$
		\While {stopping criterion not met}
                \State $Mat_{reps} \leftarrow$ Matrix of sample means and variances for all candidates
			\State $\hat{f}_j(\mbf{x}) \leftarrow$ Fit a SK metamodel to each objective $j$ using $Mat_{reps}$
                \State $Mat_{preds} \leftarrow$ Matrix of predicted means and MSEs for all candidates
			\State Compute $UCB$, $LCB$, $\widehat{UCB}$ and $\widehat{LCB}$
			\Procedure{Screening}{$UCB$, $LCB$, $\widehat{UCB}$, $\widehat{LCB}$}
				\State $\bar{S} \leftarrow$ Set of points screened out based on sample means
				\State $\widehat{S} \leftarrow$ Set of points screened out based on predicted means
				\State $\widetilde{S} = S \backslash (\bar{S} \cap \widehat{S})$ \Comment{Temporarily remove clearly inferior points}
			\EndProcedure	
                \State $\widetilde{S} = \widetilde{S} \backslash S_{max}$ \Comment Remove points that have reached $b_{max}$ 
			\State Compute $\text{EHVD}_i$ and $\text{PD}_i$, $\forall i \in \widetilde{S}$ \Comment{Eq.~\ref{eq:ehvc} and \ref{eq:pd}}
                
			\State $\widetilde{PF} \leftarrow$ Pareto front of $\widetilde{S}$ \Comment Maximization of PD and EHVD
                \State $b \leftarrow$   Vector of the number of replications allocated to each point in $\widetilde{PF}$
             \State $\widetilde{PF}_{sort} \leftarrow$ $\widetilde{PF}$ sorted in ascending order of $b$ \Comment{Prioritize allocating to solutions with low $b$}
                \While {$\widetilde{PF}_{sort}$ is not empty}
                    \For {$i \in \widetilde{PF}_{sort}$}
                        \If {$B > 0$ \AND $r_i < b_{max}$}
                            \State Allocate 1 additional replication on $\mathbf{x}_i$ \Comment Update $Mat_{reps}$
                            \State $B = B-1$
                        \ElsIf {$r_i == b_{max}$}
                            \State $S_{max} \cup \{\mbf{x}_i\}$ 
                        \Else 
                            \State \BREAK
                        \EndIf
                    \EndFor
                    \State $\widetilde{PF}_{sort} = \widetilde{PF}_{sort} \backslash S_{max}$
                \EndWhile
            \EndWhile
		\State Return the $PS$
	\end{algorithmic}
\end{algorithm}

\begin{figure}[h!]
    \centering
    \begin{subfigure}{0.49\textwidth}
        \centering
        \includegraphics[width=\linewidth]{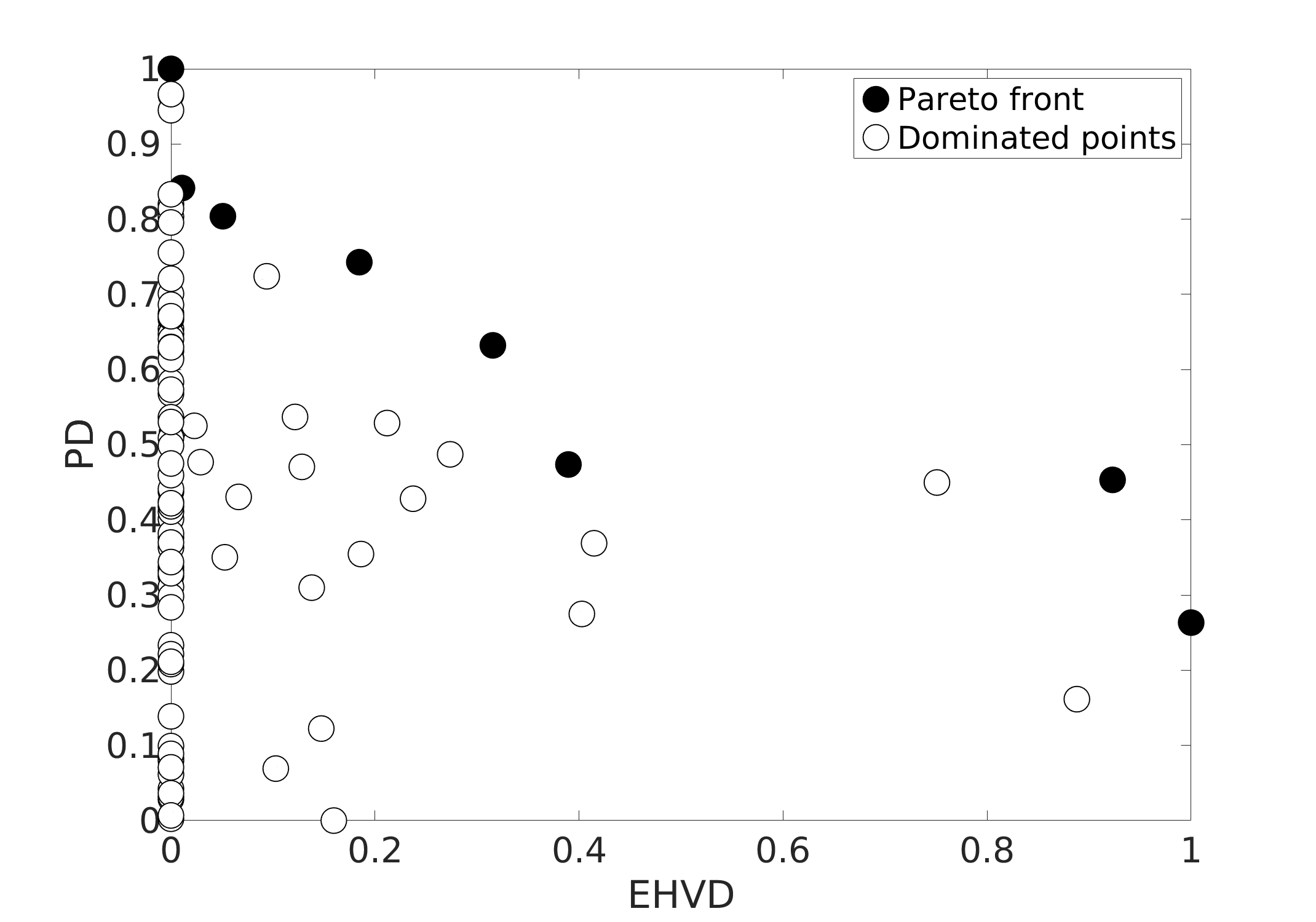}
    \end{subfigure}
    \begin{subfigure}{0.49\textwidth}
        \centering
        \includegraphics[width=\linewidth]{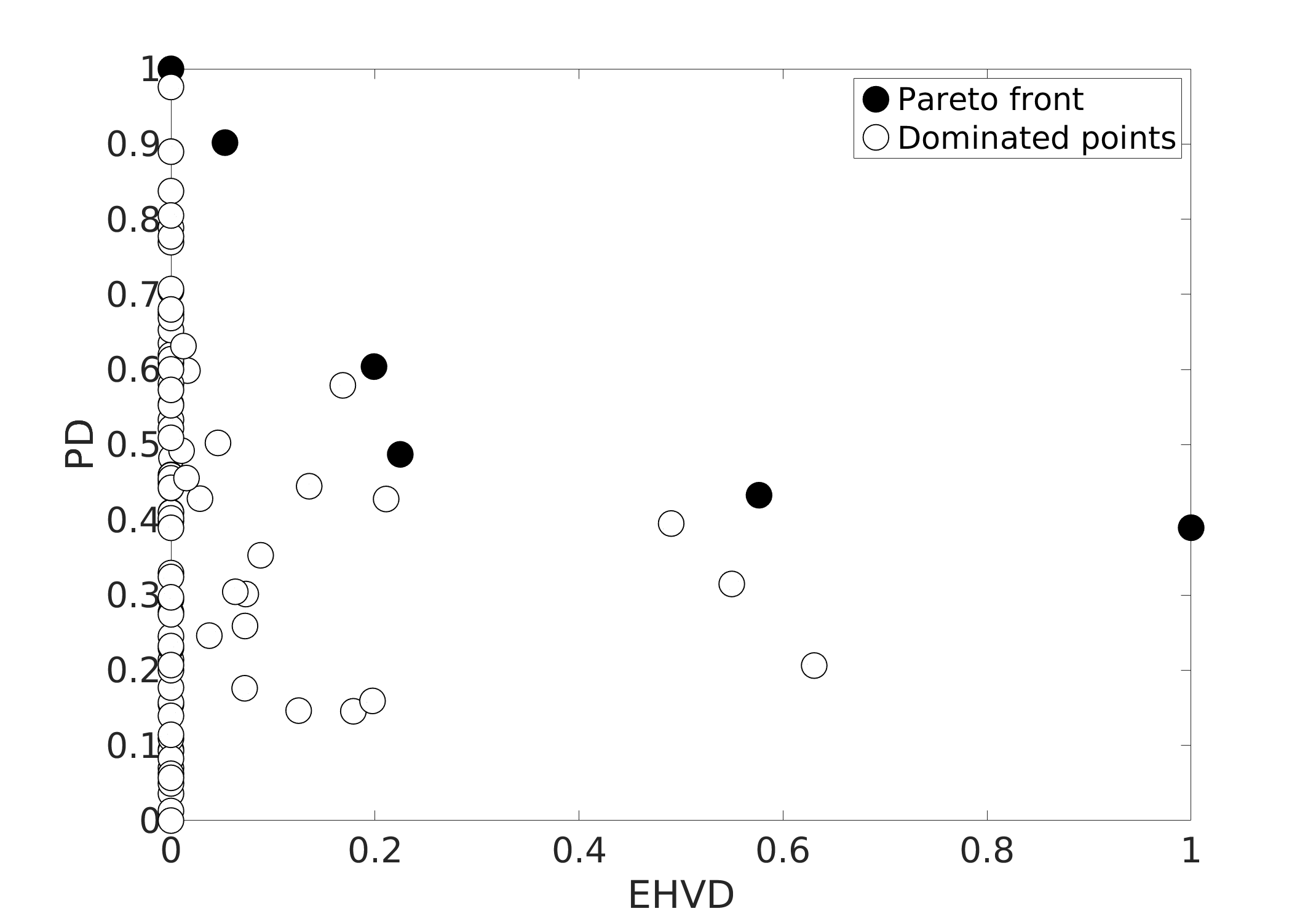}
    \end{subfigure}
\caption{Pareto front between the maximization of both criteria (normalized in the $[0,1]$ interval) at two arbitrary iterations.}
\label{fig:pfmors}
\end{figure}

After screening out a subset of clearly inferior points, the algorithm proceeds to calculate the EHVD values using Equations (\ref{sec2:eq:hv}) and (\ref{eq:ehvc}), and the PD values using Equation (\ref{eq:pd}), \emph{only} on the points that have not reached the maximum number of replications (i.e., $b_{max}$ replications). As illustrated in Figure \ref{fig:pfmors}, a Pareto front can be distinguished (denoted $\widetilde{PF}$), for which both criteria are maximized. The algorithm then iterates over this non-dominated set (sorted in ascending order according to the number of replications allocated to each point so far $r_i$), allocating one replication per point, until the budget $B$ is depleted. Note the number of points in this Pareto front can differ per iteration. 

Selecting the parameters $b_0, B$ and $b_{max}$ will depend on the specific budget and number of points in a given problem. As discussed in \cite{MOCBA}, a value of $b_0 >= 5$ is recommended to approximate the sample means and variances, and $b_{max}$ is used to prevent the algorithms to allocate an extreme number of replications to a given point. In our experiments, we use a value of $B = |S|$ (i.e, the budget per iteration equals the cardinality of the set of points). This is because the standard EQUAL allocation uniformly distributes the replication budget among \emph{all} alternatives, and thus a budget of $B = |S|$ will run 1 additional replication per point per iteration. Furthermore, a value of $b_0 = 5$ and $b_{max} = 100$ are used in our experiments for all competing algorithms. 

Recall that the two criteria serve different goals: while the EHVD allocates budget to those points that are expected to change the current HV the most, the PD focuses on improving the prediction accuracy of the performance. As it will be shown in the experimental results, allocating extra samples based on the trade-off between EHVD and PD yields better results than using either of the criteria individually. If the stopping criterion is met, the algorithm returns the \emph{predicted} Pareto set. As the proposed allocation procedure is sequential, after the budget in the current iteration has been depleted, the SK parameters are recomputed with the new (more accurate) sample means and variances to perform a new iteration. 

\section{Test problems} \label{sec:des_experiments}

In Section \ref{sec:artifproblems}, we present the implementation details of our experiments on a set of well-known analytical bi-objective test functions. These functions have different Pareto front geometries, and were implemented with varying number of decision variables, and different levels of noise. They are used to (1) evaluate the performance of the proposed allocation criteria against EQUAL allocation and the MOCBA algorithm \citep{MOCBA}, which we consider state-of-the-art; in Appendix \ref{sec:app_mocba} we provide the allocation ratios, but the full procedure is as described in \citep{chen2011stochastic}), (2) test the impact of the two screening procedures proposed, and (3) examine the impact of using the SK information instead of the sample means and variances. In addition, in Section \ref{sec:res:appl} we evaluate the performance of our proposed algorithm against EQUAL and  MOCBA on a real life  supply chain problem that we obtained from one of our industrial partners.  

All three algorithms are stopped when a predetermined number of samples have been allocated; this is a natural criterion, as in most practical settings, the sampling budget will determine the length of the run. We compare their performances by means of the \emph{accuracy of the Pareto set (APS)} metric:

\begin{equation}
    APS = 1 - \frac{MCE+MCI}{|S|}
\end{equation}

\noindent where MCI and MCE are the number of misclassification errors by inclusion and exclusion, respectively, and $|S|$ is the total number of designs. The metric is  an intuitive measure to evaluate the accuracy of the Pareto set identified by the algorithms. Evidently, it takes a value between 0 and 1: if the algorithm correctly identifies the entire true Pareto set, there is no misclassification of points, and $APS = 1$. Note that APS implicitly considers both errors to be equally important, which in the MORS context is a desirable property for algorithm evaluation. 

\subsection{Artificial test problems}\label{sec:artifproblems}
\begin{figure}[H]
	\centering
	\includegraphics[width=\textwidth]{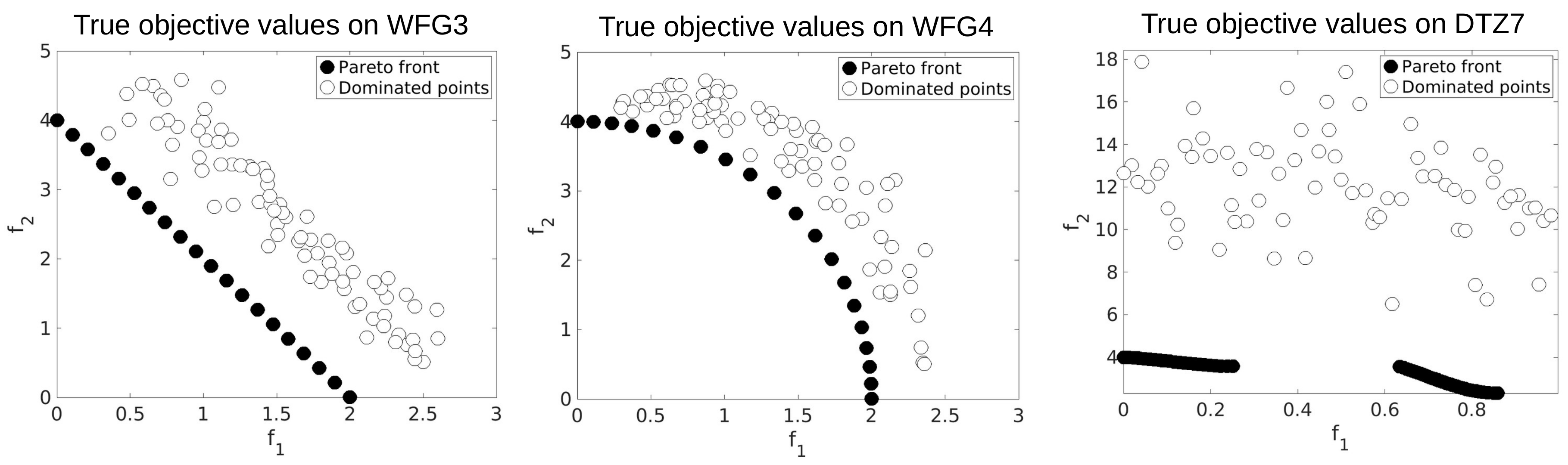}
	\caption[True Pareto fronts of the scenarios]{True Pareto front ($PF_t$, black points) and dominated points in the bi-objective scenarios.}
	\label{fig:truefronts}
\end{figure}

We choose three standard test problems from the \emph{deterministic} multi-objective literature (WFG3, WFG4 and DTLZ7) to assess the performance of the proposed procedures. The WFG test suite is state-of-the-art and allows the analyst to construct problems with any number of objectives and decision variables; features such as modality and separability can also be customized, using a set of shape and transformation functions. WFG3 is a non-separable and unimodal problem with a linear Pareto front, WFG4 is a separable and multimodal problem with a concave Pareto front, and DTLZ7 is a disconnected and multi-modal problem (see \citealt{huband2006review} for the analytical expressions and detailed characteristics of these problems). The set of points $S$ being considered is discrete and contains a fixed and known number of truly Pareto optimal points, denoted $PS_t$. Table \ref{table:scenarios} summarizes the experimental scenarios, and Figure \ref{fig:truefronts} illustrates the objective space of the test functions. 

The true objective outcomes are perturbed with heterogeneous Gaussian noise. Hence, we obtain noisy observations $\tilde{f}^k_j(\mbf{x}_i) = f_j(\mbf{x}_i)+\epsilon_j^k(\mbf{x}_i)$, with $\epsilon_j^k(\mbf{x}_i) \sim \mathcal{N}(0,\tau_j(\mbf{x}_i))$ for $j = \{1,2\}$ objectives at the $k^{th}$ replication. In accordance with the previous literature \citep{Pich13b,Jal17,Seb19}, we set the standard deviation of the noise ($\tau_j(\mbf{x})$) such that it varies linearly with respect to the objective values. The maximum and minimum values of $\tau_j(\mbf{x})$ are linked to the range of each objective value in the region of interest (i.e., $RF_j = \max_{\mbf{x} \in S} f_j(\mbf{x}) - \min_{\mbf{x} \in S} f_j(\mbf{x}), j = \{1,2\}$). In the experiments, we set the minimum noise at the individual optima (i.e. , the minima) of both objective functions. Because of the trade-off between the function values in the Pareto-optimal points, this assumption automatically leads to a trade-off in the \emph{noise} of the functions at these points (in the extremes of the bi-objective front, the noise is minimal for one objective, while maximal for the other). In that way, the linear structure ensures that the resulting R\&S problems are not trivial, as \emph{none} of the Pareto-optimal points has accurate sample means on \emph{both} objectives. Evidently, the same could have been achieved with other monotonous noise structures. As shown in Table \ref{table:scenarios}, we consider three levels of noise perturbing the responses (low, medium and high), varying between $0.001RF_j$ and $2RF_j$. 

\bgroup
\def\arraystretch{1}
\begin{table}[h]
\centering
\caption{Summary of the experimental scenarios for the artificial test functions}
\resizebox{0.85\textwidth}{!}{%
\begin{tabular}{|l|c|c|c|}
\hline
											& WFG3		 & WFG4			& DTLZ7				\\
\hline
\hline
Number of objectives $m$					&2						&2				&2			\\
\hline
Number of decision variables $d$			&5						&5				&2			\\
\hline
Number of points in $S$ ($|S|$)				&100					&100			&100		\\
\hline
Size of the true Pareto set $|PS_t|$		&20						&20				&50 		\\
\hline
Noise level									&High					&Medium			&Low		\\
\hline
Total number of iterations                  &15						&30				&30   		\\
\hline
\hline
Stochastic kriging metamodels               &\multicolumn{3}{c|}{Squared exponential kernel: Eq. \ref{eq:gauss}} \\
\hline
Low noise std. dev.                         &\multicolumn{3}{c|}{$0.001 RF_j \leq \tau_j(\mbf{x}) \leq 0.5  RF_j$} \\
\hline
Medium noise std. dev.                      &\multicolumn{3}{c|}{$0.01 RF_j \leq \tau_j(\mbf{x}) \leq 1 RF_j$} \\
\hline
High noise std. dev.                        &\multicolumn{3}{c|}{$1 RF_j \leq \tau_j(\mbf{x}) \leq 2 RF_j$} \\
\hline
\end{tabular}}
\label{table:scenarios}
\end{table}
\egroup

\subsection{Practical application from industry} \label{sec:res:appl}

We also test the competing algorithms on a real-life multi-period supply chain problem, obtained from one of our industrial partners in the chemical process industry. The company has several plant sites, each of them equipped with several processing units for production. Each of these units has unique processing capabilities and is responsible for a different category of products. Each category consists of products showing similarities in their physical and chemical properties. The customers are separated into sales regions, and the demand forecasts per period (for all products) are generated at the level of these sales regions. The demand uncertainty for each product is modeled using a normal distribution, where the mean of this distribution is given by the forecast, and the standard deviation is approximated using historical data. Transportation is assumed to always be available, on each route from the sites to the regions. This problem setting has already been used as a real-life case in other articles (e.g., \citealt{jung2004simulation}), but then as a single-objective problem with a stochastic constraint (i.e., minimize the cost function subject to a service level constraint).  

The company needs to decide the number of units produced per week, for each product and each facility. Any excess product is stored in a warehouse next to the production facility. Once demand in each sales region is known, it is satisfied by the available supply from the facility closest to the sales region (i.e., pairs of sales region and facility are sorted in increasing distance, and demand is satisfied in this order). Any demand that cannot be satisfied by the company is outsourced and satisfied by a third party at a higher cost. There are two objectives: minimizing total cost (consisting of production cost, outsourcing cost, transportation cost, and inventory holding cost), and maximize customer service level which is defined as the percentage of periods that the company does \emph{not} need outsourcing to meet customer demand. In the example below we use 3 production facilities and 5 products, so a total of 15 decision variables. Based on \cite{jung2004simulation}, we provide a more detailed formulation of the problem in Appendix \ref{sec:app_prac}. Each simulation period is a week and the simulation runs over 1 year (i.e., 52 periods). Evidently, the demand from the customers is uncertain, and we consider different levels of uncertainty by varying the coefficient of variation of the demand, denoted $\chi$ (see Appendix \ref{sec:app_prac}). We evaluate 4 different scenarios: $\chi = 0.1, 0.3, 0.6 \text{ and } 0.9$. Table \ref{table:scenarios2} summarizes the parameters of the scenarios. 

\bgroup
\def\arraystretch{1}
\begin{table}[h]
\centering
\caption{Summary of the experimental scenarios for the practical problem}
\resizebox{0.85\textwidth}{!}{%
\begin{tabular}{|l|c|c|c|c|}
\hline
											& Scenario 1      & Scenario 2			& Scenario 3	& Scenario 4\\
\hline
\hline
Number of points in $S$ ($|S|$)				&125					&96	    		&93	              &99     	\\
\hline
Size of the true Pareto set $|PS_t|$		&33						&32				&22               &25		\\
\hline
Coefficient of variation of demand									&$\chi = 0.1$			&$\chi = 0.3$	&$\chi = 0.6$	  &$\chi = 0.9$\\
\hline
\end{tabular}}
\label{table:scenarios2}
\end{table}
\egroup

\begin{figure}[h!]
    \centering
    \begin{subfigure}{0.49\textwidth}
        \centering
        \includegraphics[width=\linewidth]{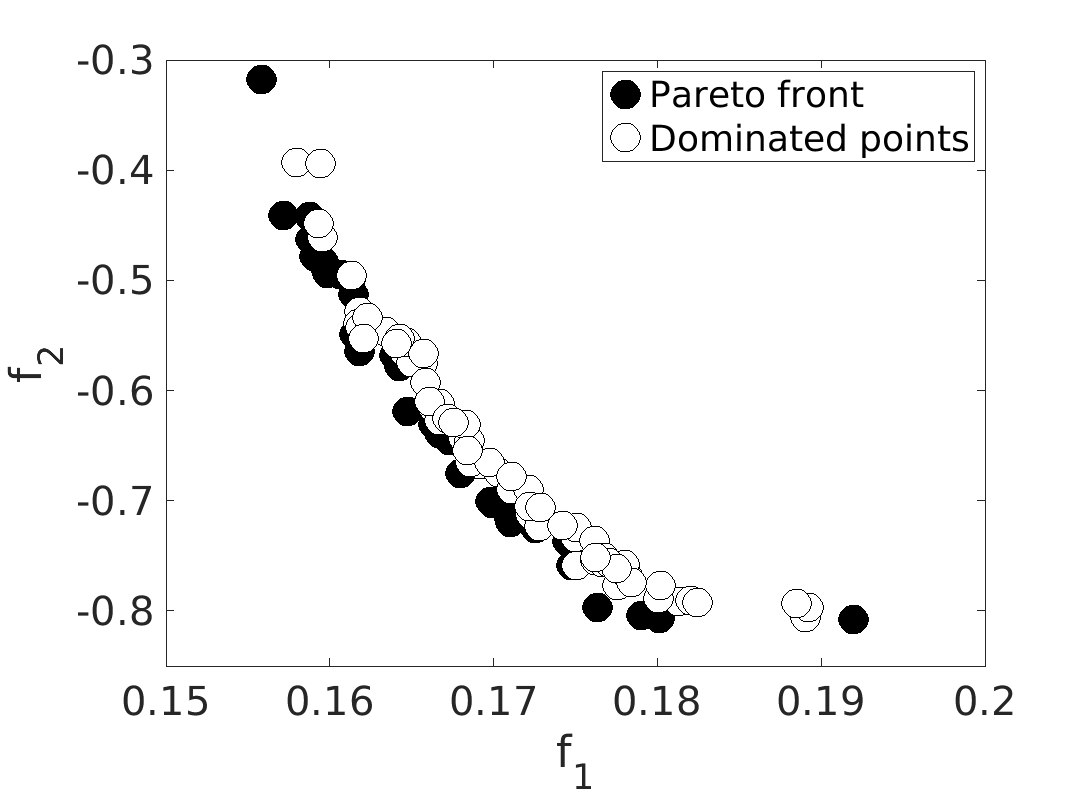}
    \end{subfigure}
    \begin{subfigure}{0.49\textwidth}
        \centering
        \includegraphics[width=\linewidth]{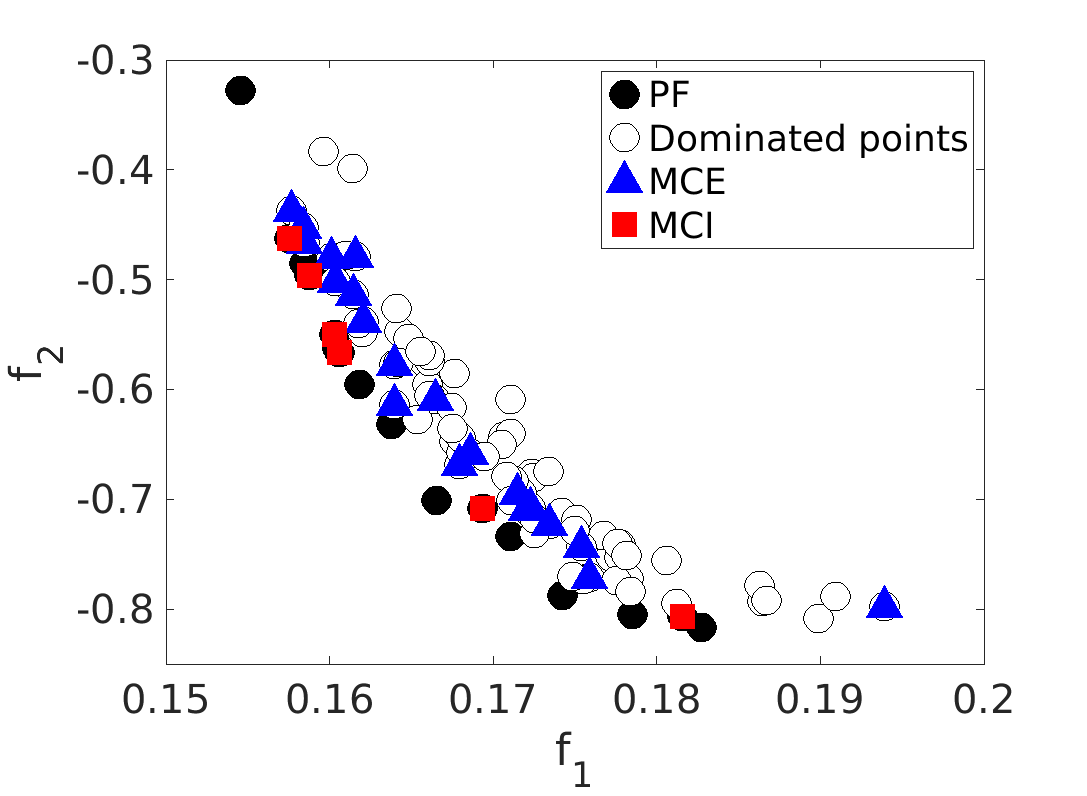}
    \end{subfigure}
\caption{Pareto front for the simultaneous minimization of cost (denoted $f_1$) and negative service level (denoted $f_2$). The left panel shows the approximation of the true performance, and the right panel the misclassification errors at an arbitrary iteration.} 
\label{fig:app}
\end{figure}

\section{Results} \label{sec:experiments}

We first evaluate the performance of the proposed screening procedures in Section~\ref{sec:res:screening}. In Section \ref{sec:res:main}, we compare the SK-MORS algorithm against EQUAL allocation and the MOCBA algorithm. Furthermore, we show that resampling points based on the combination of EHVD and PD outperforms a resampling approach based on either of these criteria in isolation. In Section \ref{sec:res:upgrade}, we analyze the benefit of using the SK information instead of the sample means in the EQUAL allocation and MOCBA algorithms. Finally, in Section \ref{sec:res:application} we evaluate the performance on the practical problem. In all cases, we run the algorithms 30 times on the same instance (i.e., 30 macroreplications); in the figures, we report the average APS value obtained at each iteration of the algorithms, along with the 95\% confidence intervals. Table \ref{table:algos} summarizes the algorithms evaluated.

\bgroup
\def\arraystretch{1.2}
\begin{table}[h]
\centering
\caption{Overview of the different algorithms tested.}
\resizebox{\textwidth}{!}{%
\begin{tabular}{ |c|l| }
\hline
Algorithm & Description\\
\hline
\hline
SKMORS$_{none}$ & SK-MORS algorithm without using a screening heuristic. \\
\hline
SKMORS$_{box}$ & SK-MORS algorithm using the \textsc{Screening Box} heuristic (Algorithm \ref{alg:screening1}). \\
\hline
SKMORS$_{band}$ & SK-MORS algorithm using the \textsc{Screening Band} heuristic (Algorithm \ref{alg:screening2}). \\
\hline
SKMORS-PD & SK-MORS algorithm using only the PD criterion.  \\
\hline
SKMORS-HV & SK-MORS algorithm using only the EHVD criterion. \\
\hline
EQUAL & Allocate replications uniformly to all sampled points. \\
\hline
EQUAL-SKi & EQUAL allocation using the SK predictions for identification. \\
\hline
MOCBA & Multiobjective Optimal Computing Budget Allocation algorithm. \\
\hline
MOCBA-SK & MOCBA algorithm using the SK predictions for allocation and identification. \\
\hline
MOCBA-SKi & MOCBA algorithm using the SK predictions only for identification. \\
\hline
\end{tabular}}
\label{table:algos}
\end{table}
\egroup

\subsection{Impact of the screening procedures} \label{sec:res:screening}

\begin{figure}[h!]\centering
\subfloat[Average number of points considered at each iteration for scenario WFG3 (high noise)]{\label{b}\includegraphics[width=.5\linewidth]{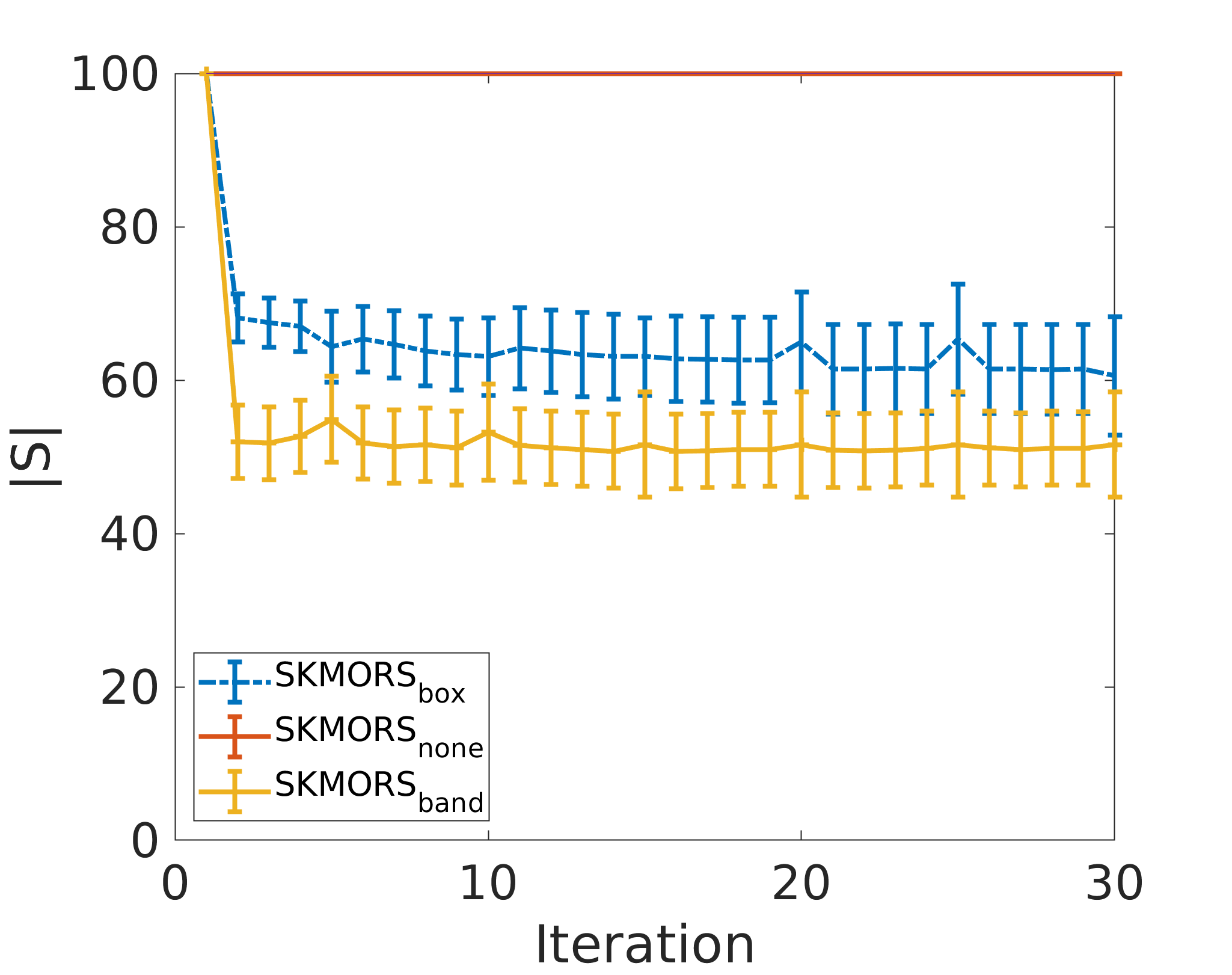}}\hfill
\subfloat[Performance on WFG3 (high noise)]{\label{a}\includegraphics[width=.5\linewidth]{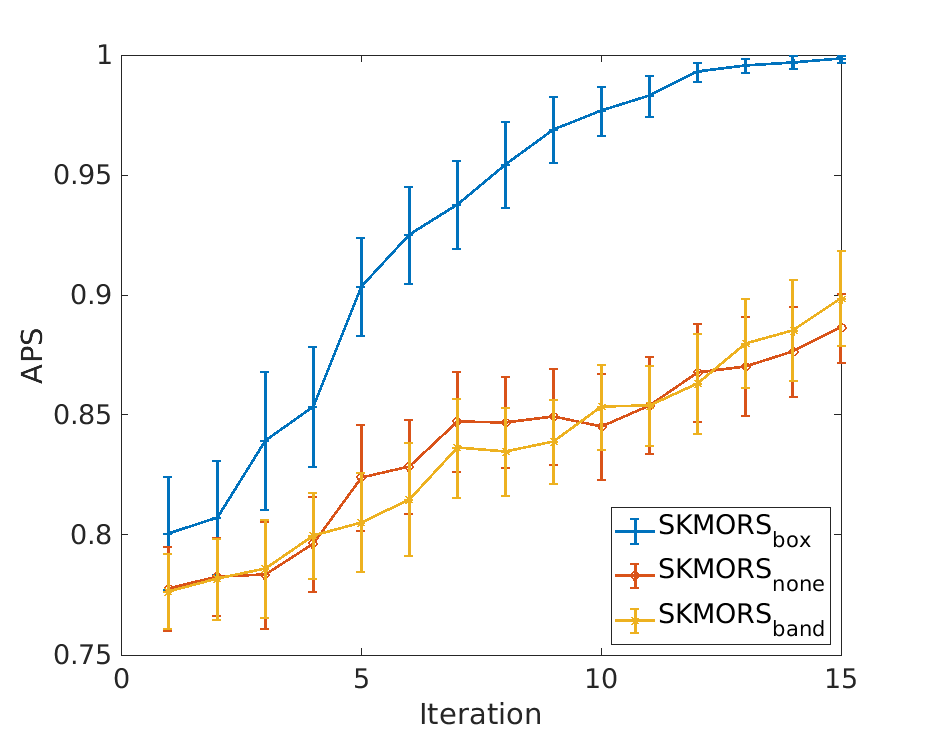}}\par
\subfloat[Average number of points considered at each iteration for scenario WFG4 (medium noise)]{\label{b}\includegraphics[width=.5\linewidth]{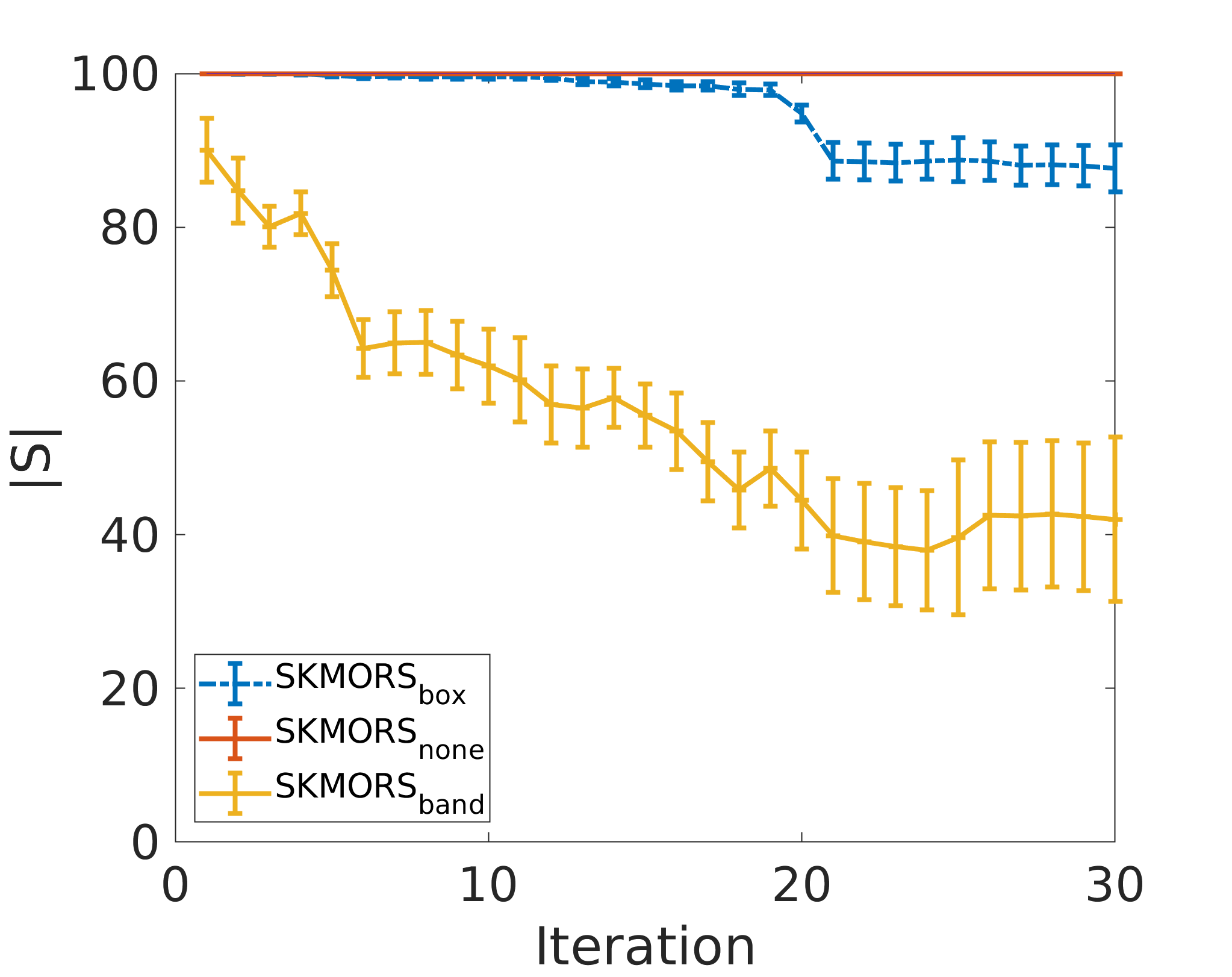}}\hfill
\subfloat[Performance on WFG4 (medium noise)]{\label{a}\includegraphics[width=.5\linewidth]{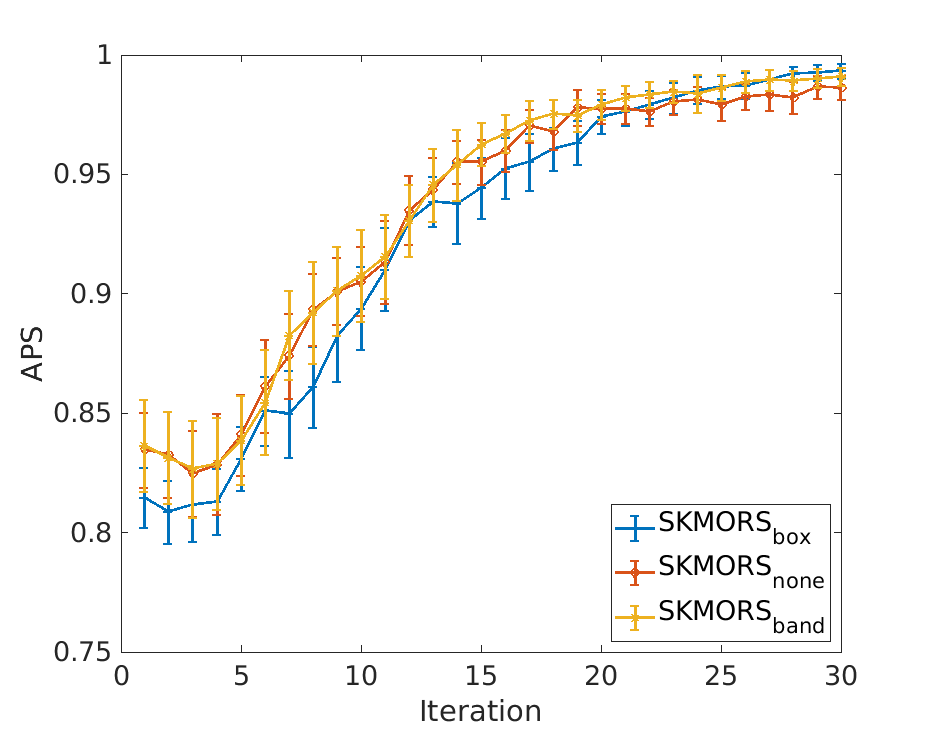}}\par
\subfloat[Average number of points considered at each iteration for scenario DTLZ7 (low noise)]{\label{b}\includegraphics[width=.5\linewidth]{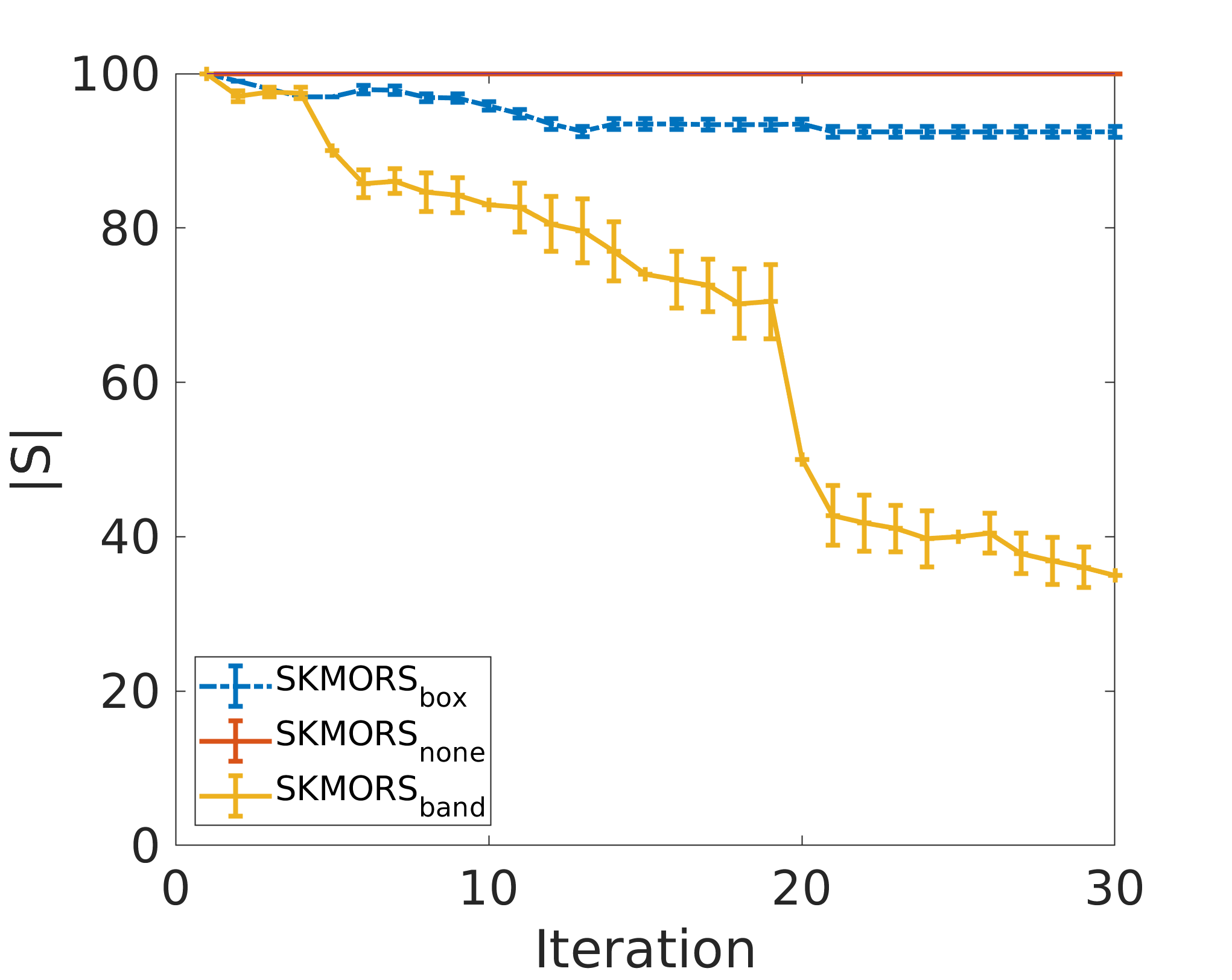}}\hfill
\subfloat[Performance on DTLZ7 (low noise)]{\label{a}\includegraphics[width=.5\linewidth]{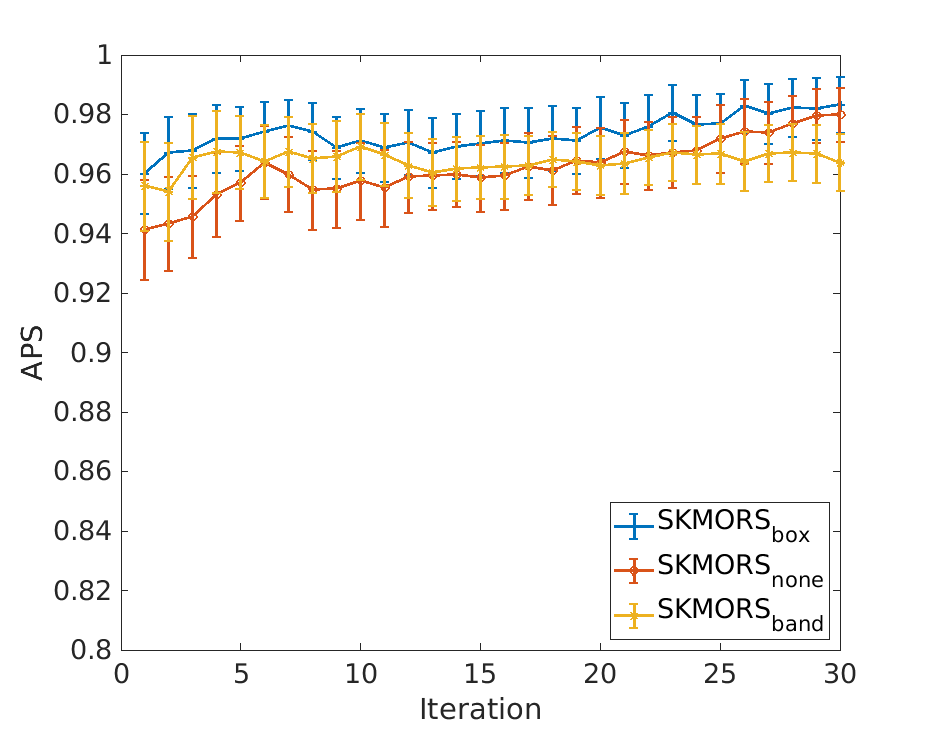}}\par
\caption{Performance of the proposed screening procedures.}
\label{fig:res2}
\end{figure}

The impact of the proposed screening procedures is shown in Figure \ref{fig:res2}; the metric $|S|$ shows the number of points that remain in the set after screening, at a given iteration. Both screening procedures succeed in sequentially reducing the number of candidate points considered as the observed performance becomes more accurate (see (a), (c) and (e) in Figure \ref{fig:res2}); evidently, the biggest reduction is obtained with \textsc{Screening Band}. Such reduction significantly reduces the computational cost of executing the hypervolume calculations in Eq. \ref{eq:ehvc} for every candidate point at each iteration. 

As clear from Figures \ref{fig:res2}(b), (d) and (f), the noise levels for the different scenarios have an impact on the performance of the proposed algorithm w.r.t. the the convergence to the true front. While for low and medium levels of noise (i.e., scenarios DTLZ7 and WFG4 respectively) no particular differences are observed between both proposed methods, for a high level of noise (i.e., scenario WFG3) the performance of \textsc{Screening Band} clearly worsens. This is not entirely surprising, as filtering out some points may result in leaving out truly non-dominated points (i.e., MCE points) in one or more iterations, especially for the \textsc{Band} procedure. Moreover, in Figure  \ref{fig:res2}(e), we observe that the BAND procedure screens out a very large number of points, including some of the MCE errors. This happens because the noise is low and the dominated points are relatively far from the non-dominated points. \textsc{Screening Box} on the other hand, shows consistent performance; thus in further sections we use SK-MORS$_{box}$ as the default proposed algorithm.

\subsection{Proposed allocation criteria} \label{sec:res:main}

\begin{figure}[h!]\centering
\subfloat[Performance on WFG3 (high noise)]{\label{a}\includegraphics[width=.5\linewidth]{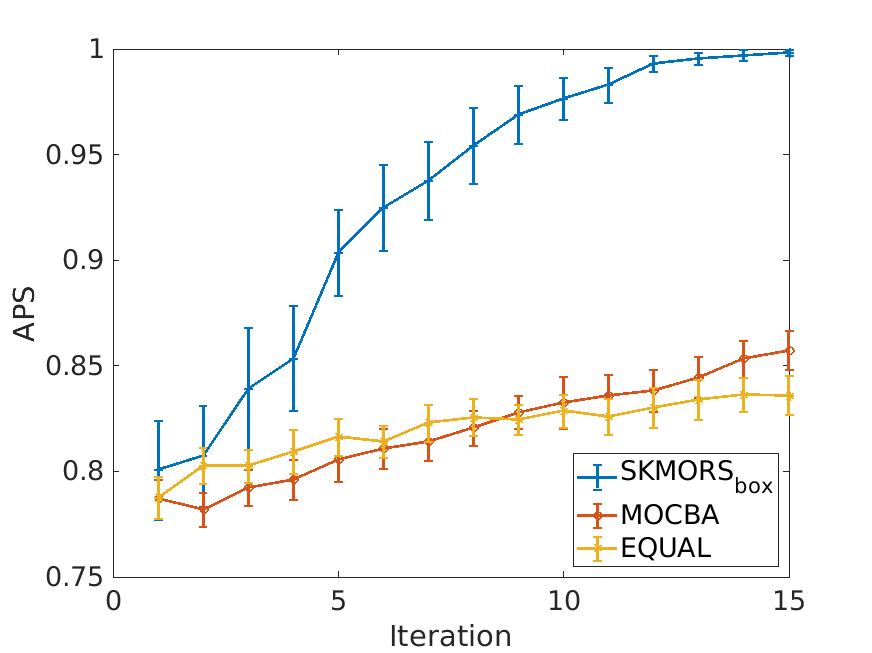}}\hfill
\subfloat[Performance on WFG4 (medium noise)]{\label{b}\includegraphics[width=.5\linewidth]{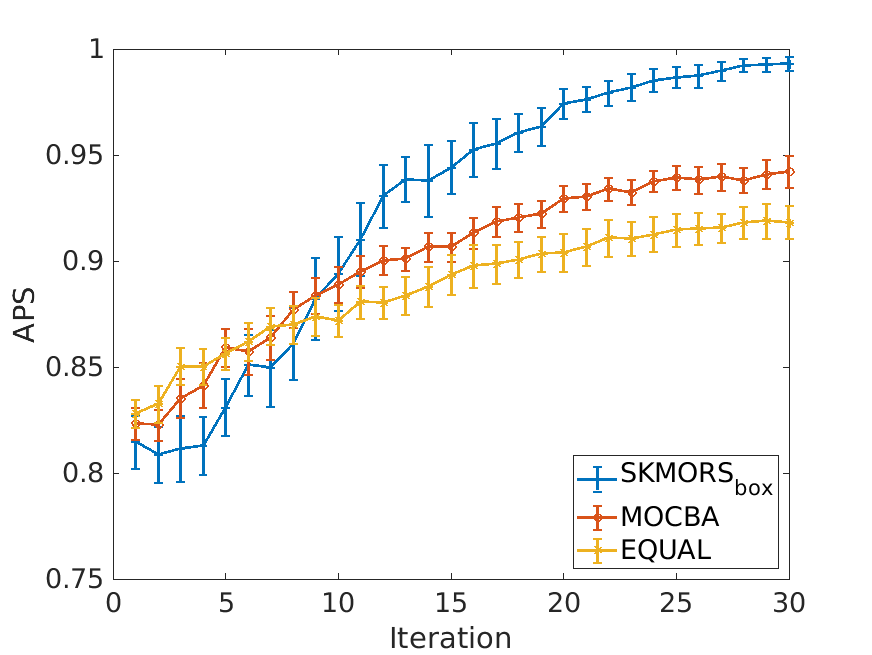}}\par 
\subfloat[Performance on DTLZ7 (low noise)]{\label{c}\includegraphics[width=.5\linewidth]{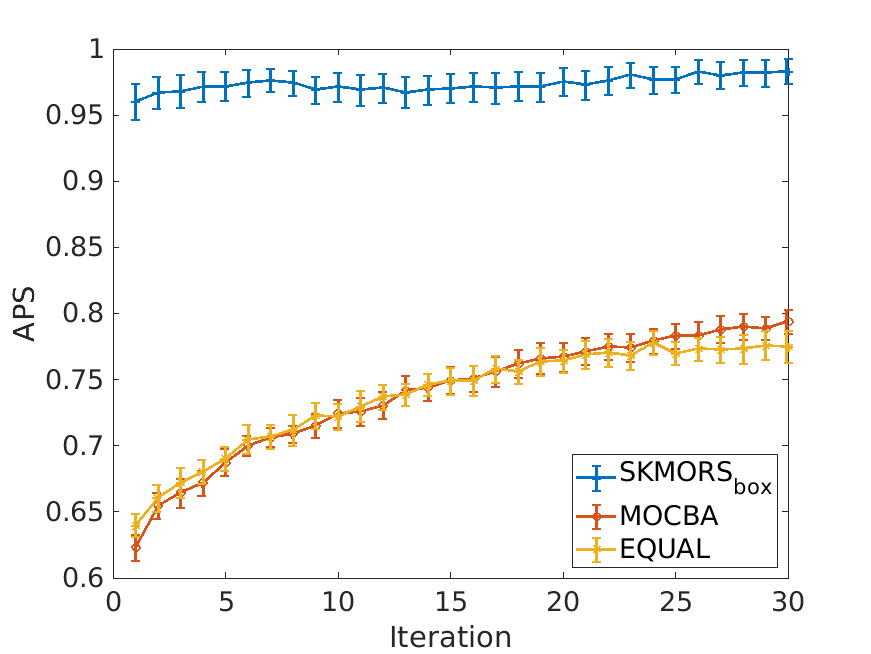  }}\hfill
\caption{Convergence to the true Pareto front: the proposed algorithm versus standard EQUAL allocation, and the state-of-the-art MOCBA procedure, for the 3 test problems considered.}
\label{fig:res1}
\end{figure}

Fig \ref{fig:res1} clearly shows that, for all test problems considered, our algorithm converges much faster towards the correct identification of the true Pareto front than EQUAL or MOCBA. Note that the superiority of MOCBA vs. EQUAL is not entirely clear for all scenarios (e.g., Figures \ref{fig:res1}(a) and \ref{fig:res1}(c)), as in these scenarios MOCBA and EQUAL require a much larger budget to converge. On the other hand, the proposed algorithm shows consistent performance. It is also remarkable that in scenario DTLZ7 the SK predictions provide clearly superior estimates already from the beginning of the procedure, which we attribute to the low level of noise compared to the other scenarios. 

\begin{figure}[h!]
	\centering
	\includegraphics[width=0.65\textwidth]{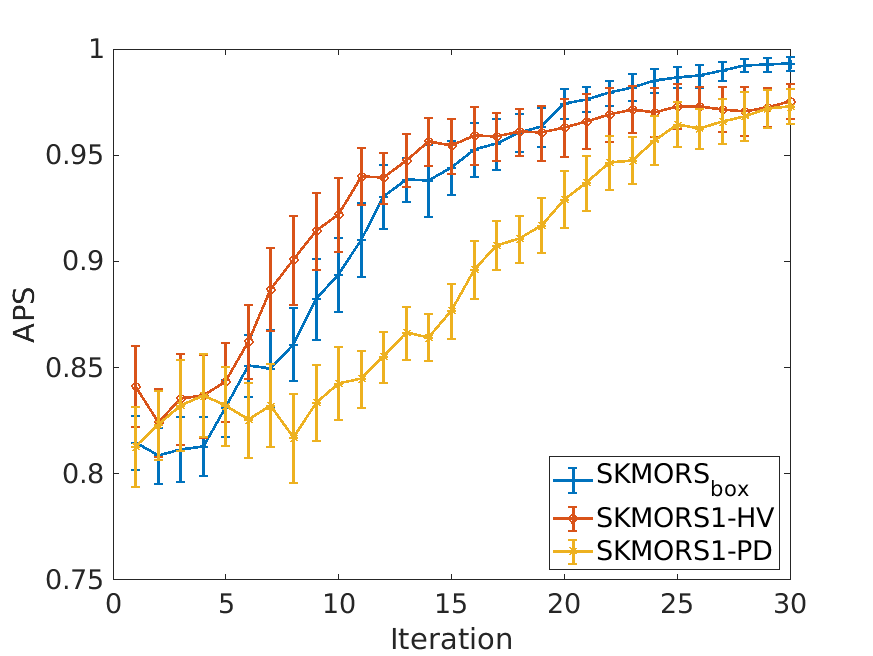 }
	\caption{Individual performance of the proposed resampling criteria on WFG4 (medium noise).}
	\label{fig:res4}
\end{figure}

Figure \ref{fig:res4} shows that the convergence is faster when both criteria are combined, as opposed to using the PD and EHVD separately (see Figure \ref{fig:pfmors}). This is expected, given that one criterion complements the limitations of the other. An advantage of the PD criterion is its very low computational cost, although the EHVD is shown to be more efficient in convergence (also observed in further experiments with other scenarios). 

\subsection{Impact of stochastic kriging} \label{sec:res:upgrade}

\begin{figure}[H]
	\centering
	\includegraphics[width=0.65\textwidth]{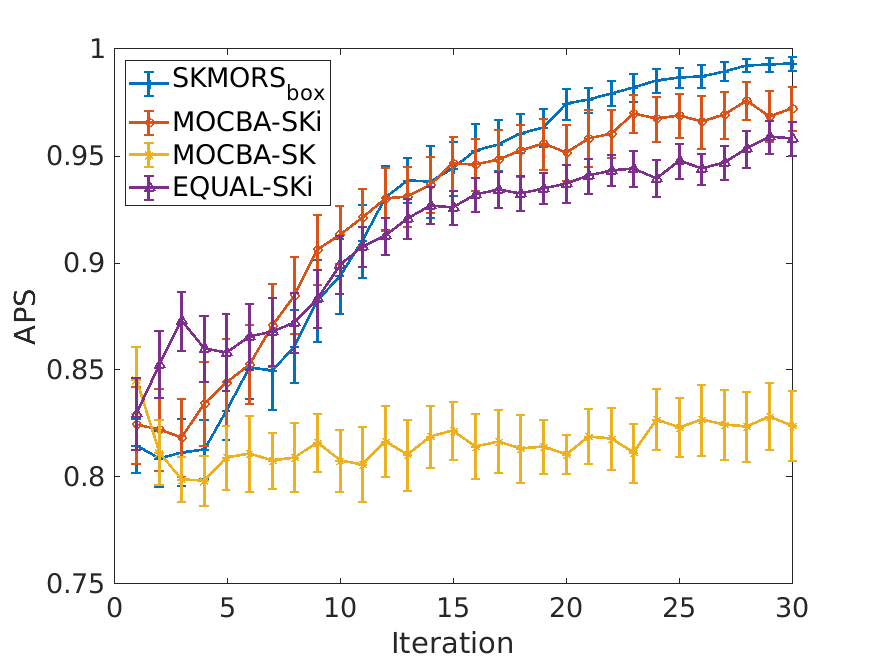}
	\caption[Benefiting]{Performance on WFG4 (medium noise).}
	\label{fig:res3}
\end{figure}

As evident from Figure \ref{fig:res3}, feeding SK information to the identification phase of EQUAL and MOCBA (i.e., using SK predictions only when computing the misclassification errors), leads to a drastic improvement in both algorithms (referred to as EQUAL-SKi and MOCBA-SKi respectively); yet, they still show worse performance than the proposed method. The bad performance of MOCBA-SK (i.e., feeding SK information to the MOCBA algorithm for allocation and identification) is somewhat surprising; however, as the sample variance and prediction MSE don't reflect the same information (i.e., sampling variability and prediction error respectively). Overall, it is clear that the efficient identification of the solutions with the true best expected performance is facilitated by exploiting the stochastic kriging information (as opposed to relying on the sample means).

\subsection{Performance on the supply chain problem} \label{sec:res:application}

\begin{figure}[h]\centering
\subfloat[Performance on Scenario 1]{\label{a}\includegraphics[width=.5\linewidth]{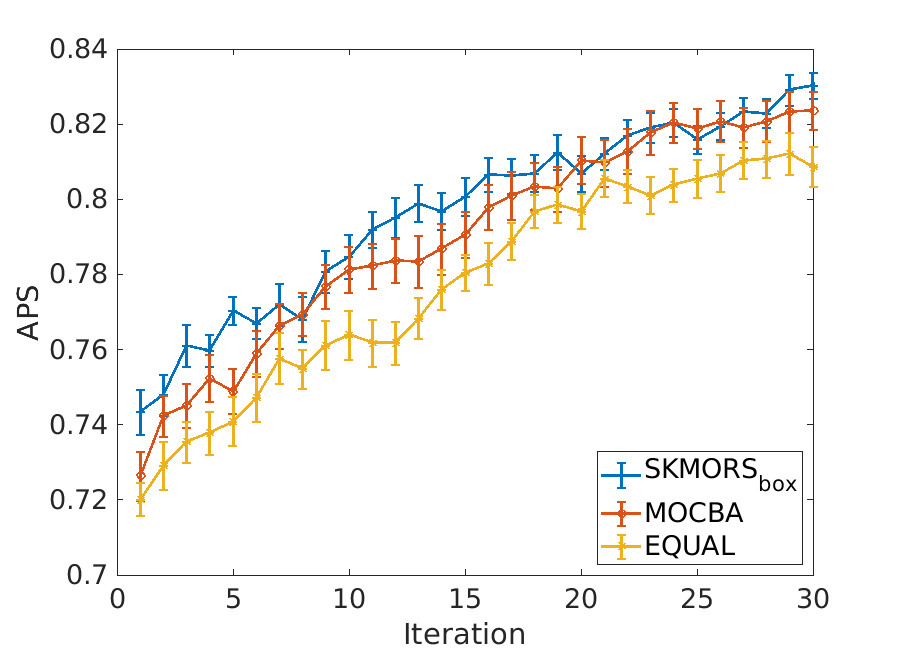}}\hfill
\subfloat[Performance on Scenario 2]{\label{b}\includegraphics[width=.5\linewidth]{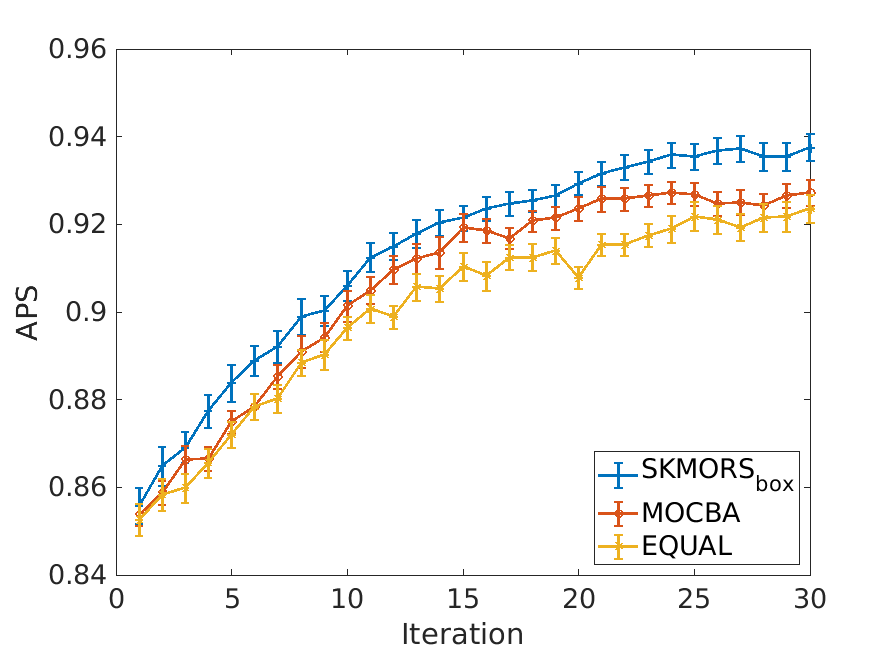}}\par 
\subfloat[Performance on Scenario 3]{\label{c}\includegraphics[width=.5\linewidth]{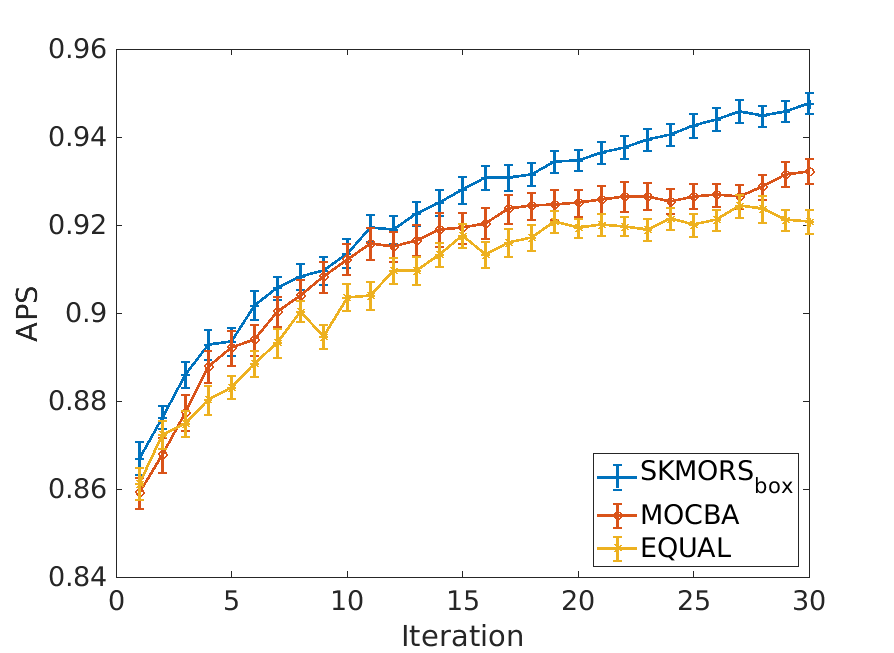}}\hfill
\subfloat[Performance on Scenario 4]{\label{b}\includegraphics[width=.5\linewidth]{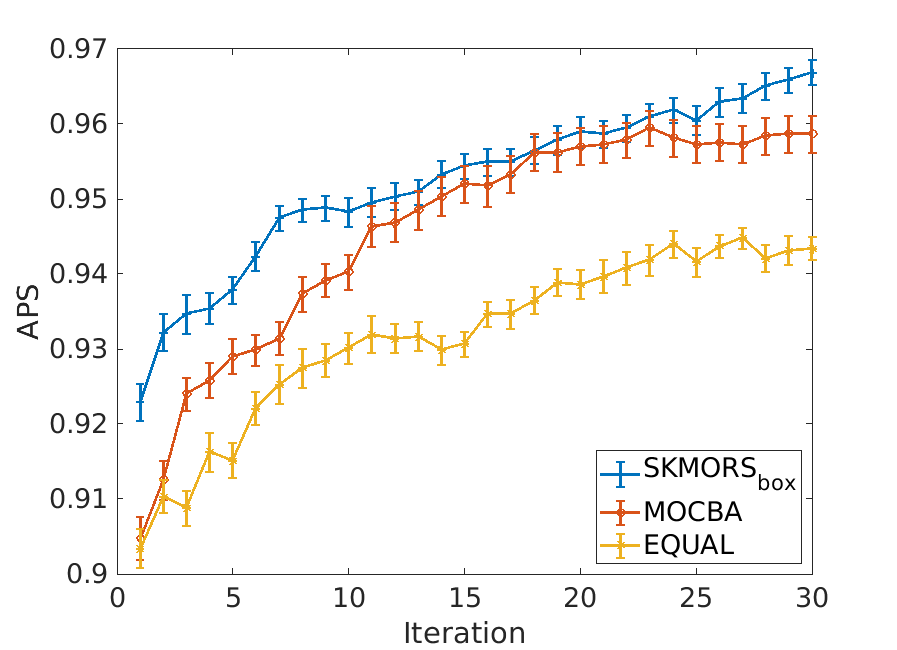}}\par 
\caption{Performance of the proposed algorithm versus standard EQUAL and MOCBA, for the 4 scenarios considered.}
\label{fig:res_app}
\end{figure}

Figure \ref{fig:res_app} shows that the proposed method remains superior to MOCBA and EQUAL; with increasing noise levels, the superiority becomes more evident. This result is analogous to those observed with analytic test functions. A relevant observation that is not visible from the figure, all algorithms struggle to converge to zero errors on this problem, even with very large budgets of 100 or more iterations, because of several solutions with only marginally different performance (conversely, see the results in Figure \ref{fig:res1}). Indeed, in practice we will often encounter several solutions whose performance is marginally different, and thus also incorporating a \emph{multiobjective indifference zone} procedure can be beneficial. Yet again, as discussed in Section \ref{sec:lr}, multiobjecive IZ procedures remain very scarce in the literature. Finally, in Table \ref{table:runningtime} we report the overhead running time per iteration for all the algorithms. Evidently, fitting the metamodels and quantifying the acquisition functions implies a significant computational effort, especially in this case with 15 input dimensions). In fact, 15 input dimensions is already close to the limit for Gaussian Process regressors \cite{Rasm06,Kleijnen15}; higher dimensions would require implementing a dimensionality reduction technique (see e.g., \citep{binois2022survey}], which in turn incorporates yet another source of error. 

\bgroup
\def\arraystretch{1}
\begin{table}[H]
\centering
\caption{Average (std. dev.) running time per macroreplication in seconds for the practical application (including the simulation evaluations).}
\resizebox{0.75\textwidth}{!}{%
\begin{tabular}{ |c|c|c|c|c| }
\hline
SK-MORS$_{none}$ & SK-MORS$_{box}$ & SK-MORS$_{band}$ & MOCBA & EQUAL\\
\hline
\hline
57.77 (12.01) & 52.44 (9.97) & 31.21 (8.40) & 7.73 (0.44) & 5.77 (0.21)\\
\hline
\end{tabular}}
\label{table:runningtime}
\end{table}
\egroup

\section{Conclusions and future research} \label{sec:conclusion}

This paper proposed a bi-objective ranking and selection technique that can deal with heteroscedastic noise and exploit spatial correlations between the alternatives by using stochastic kriging metamodels. The proposed sampling criteria are able to allocate samples efficiently in view of minimizing both misclassification errors. A high value for the EHVD criterion at a given point indicates that extra replications at that point are expected to generate a drastic change to the estimated front, implying that it is interesting for the decision maker to spend additional replication budget there. This criterion tends to be highest for points that are expected to be on the true front. The PD on the other hand, focuses on improving the accuracy of all the alternatives in objective space, and explicitly neglects the dominance relationships between the alternatives. The PD has the additional advantage of being significantly cheaper to compute. By selecting the points that are on the Pareto front when simultaneously maximizing both criteria, we obtain an algorithm that efficiently allocates the available replication budget in view of reducing misclassification errors. 

To reduce the computational effort, we proposed two screening procedures: as shown, the use of these procedures did not significantly impact the performance of the algorithm in terms of convergence to the true front with low and medium noise levels. The performance of the \textsc{Screening Band} procedure was observed to deteriorate with higher noise levels, as it has a higher risk of filtering out truly non-dominated points (i.e., MCE points). The choice of which one to use in practice will depend on the specific problem at hand. For example, if detecting small differences in performance for the observed non-dominated solutions is not crucial for the decision-maker, using \textsc{Screening Band} can be helpful to reduce the computational overhead significantly, while still considering well-performing solutions. Moreover, both procedures may be combined, first using \textsc{Screening Box} and then switching to \textsc{Screening Band} in later iterations. 

Especially in settings with high noise, the sample variance can be inaccurate with respect to the true variance if only a small number of replications is taken. Yet, the results obtained on the real-life supply chain problem illustrate that the proposed algorithm consistently performs well, even if the (endogenous) noise on the objective values is increased by manipulating the coefficient of variation of the demand. We are quite confident that the assumption of linear noise structure in the analytical test problem does not significantly influence our results; yet, other noise structures may be tested. This poses an interesting avenue for further work.

Extending the algorithm to more objectives seems straightforward, as we can build metamodels for as many objectives as is required. However, the computational cost for the hypervolume computations in Equation \ref{eq:ehvc} would increase substantially. Higher-dimensional objective spaces are an open question in MORS research, as MORS procedures tend to invest an extremely large replication budget to differentiate solutions that are marginally different \citep{Feldman18,cooper2020pymoso}, especially when the number of alternatives are in the hundreds or thousands (as with many objectives). In this context, analyzing the parallelization of the resampling criteria and/or resampling in batches of solutions (e.g., by computing the batch of solutions that jointly maximize the EHVD) would evidently be valuable. We thus believe that a thorough study with 3 or more objectives is non-trivial, and poses a major challenge to the MORS community as well.

\bibliography{references}
\newpage 

\begin{appendices}
\section{MOCBA allocation rules} \label{sec:app_mocba}

The allocation rules for the simplified MOCBA algorithm \cite{chen2011stochastic} are summarized below:
\begin{itemize}[label={}]
\item $\bar{f}_{ij}:$ The averaged observed performance of design $i$ for objective $j$ after a certain number of replications.
\item $p_i:$ The design that dominates design $i$ with the highest probability.
\item $j^i_{p_i}:$ The objective $j$ of $p_i$ that dominates the corresponding objective of design $i$ with the lowest probability.
\item $\tau_{ij}:$ The observed intrinsic standard deviation of design $i$ for objective $j$ after a certain number of replications. 
\item $\alpha_i:$ The budget allocation for design $i$. 
\item $S$: The entire set of sampled points. 
\item $S_A$: The subset of designs in $S$ labeled as being dominated.
\item $S_B$: The subset of designs in $S$ labeled as being non-dominated. 
\end{itemize}

For any given design $g,h \in S_A$ and $d \in S_B$:
\begin{align}
\alpha_h &= \left(\frac{\tau_{hj^h_{p_h}} / \delta_{hp_hj^h_{p_h}}}{\tau_{gj^g_{p_g}} / \delta_{gp_gj^g_{p_g}}}\right)^2 \label{eq:alloc1}\\
\alpha_d &= \sqrt{\sum_{h \in D_d}\frac{\tau^2_{dj_d^h}}{\tau^2_{hj_d^h}}\alpha_h^2} \label{eq:alloc2}
\end{align}
where
\begin{align}
\delta_{ipj} &= \bar{f}_{pj} - \bar{f}_{ij} \label{eq:mocba1} \\
j^i_p &= \argmin_{j \in \{1,...,m\}} P(\bar{X}_{pj} \leq \bar{X}_{ij}) = \argmax_{j \in \{1,...,m\}} \frac{\delta_{ipj}|\delta_{ipj}|}{\tau^2_{ij}+\tau^2_{pj}}\\
p_i &= \argmax_{\substack{p \in S \\ p \neq i}} \prod_{j = 1}^m P(\bar{f}_{pj} \leq \bar{f}_{ij}) = \argmin_{\substack{p \in S \\ p \neq i}} \frac{\delta_{ipj_p^i}|\delta_ipj_p^i|}{\tau^2_{ij_p^i}+\tau^2_{pj_p^i}}\\
S_A &= \left\lbrace h|h \in S,\frac{\delta^2_{hp_hj^h_{p_h}}}{\tau^2_{hj^h_{p_h}}+\tau^2_{p_hj^h_{p_h}}} < \min_{i \in D_h}\frac{\delta^2_{ihj_h^i}}{\tau^2_{ij_h^i}+\tau^2_{hj_h^i}} \right\rbrace \label{SA}\\
S_B &	= S \setminus S_A \label{SB} \\
D_h &= \left\lbrace i|i \in S, p_i = h \right\rbrace \\
D_d &= \left\lbrace h|h \in S_A, p_h = d \right\rbrace \label{eq:mocba7}
\end{align}

\section{Formulation of the practical problem} \label{sec:app_prac}
We define the following: 
\begin{itemize}[label={}]
    \item $P_{ijst}$: production amount of product $i$ on processing unit $j$ at facility $s$ in time period $t$, this is a decision variable. 
    \item $C_{ijs}$: unit production cost of product $i$ on processing unit $j$ at facility $s$. 
    \item $S_{isct}$: supply of product $i$ from facility $s$ to sales region $c$ in time period $t$, this is a decision variable.
    \item $t_{sc}$: unit transportation cost to move a unit of product from facility $s$ to sales region $c$.
    \item $\Gamma_{ict}$: amount of product $i$ to customer $c$ in time period $t$ that must be outsourced due to insufficient supply at company’s facilities.
    \item $\xi_{ic}$: unit price of outsourcing for product $i$ demanded in sales region $c$.
    \item $I_{ist}$: inventory level for product $i$ at the end of time period $t$ at facility $s$.
    \item $h_{ist}$: inventory cost for holding a unit of product $i$ in facility $s$ for the duration of time period $t$.
    \item $R_{ijst}$: effective rate for product $i$ on processing unit $j$ at facility $s$ in time period $t$.
    \item $RL_{ijst}$: run length of product $i$ on processing unit $j$ at facility $s$ in time period $t$.
    \item $H_{jst}$: amount of time available for production on process $j$ at facility $s$ during time period $t$.
    \item $\omega_{ict}$: uncertain demand for product $i$ from customer $c$ at time period $t$.
    \item $Z_{ict}$: binary variable that takes value 1 if demand in sales region $c$ in time period $t$ is met through
    production at one of the facilities, and 0 if the demand is met through outsourced production.
    \item $M_b$: maximum order size for a sales region in a single time period.
    \item $M_s$: minimum order size for a sales region in a single time period.
    \item $\chi$: coefficient of variation of the demand.
    \item $N_c$ is the number of sales regions.
    \item $N_t$ is the number of time periods.
\end{itemize}
\noindent The minimization of the cost and maximization of service level is formulated as
\begin{align}
    \min \hspace{1em} &\sum_{i,j,s,t} C_{ijs} P_{ijst} +  \sum_{i,s,c,t} t_{sc} S_{isct} + \sum_{i,c,t} \xi_{ic} \Gamma_{ict} + \sum_{i,s,t} h_{ist} I_{ist}  \label{eq:pracobj1} \\
    \max \hspace{1em} &\frac{1}{N_c N_t} \sum_{c,t}Z_{ict} \label{eq:pracobj2} 
\end{align}
\noindent subject to:
\begin{align*}
    &P_{ijst}, S_{isct}, \Gamma_{ict}, I_{ist} \geq 0 \hspace{13em} \text{non-negativity constraint} \\[18pt] 
    &P_{ijst} = R_{ijs}(RL_{ijst})                                                      \\
    &P_{ijst} \leq R_{ijs}H_{jst} \hspace{16em} \text{production capacity constraints}               \\[18pt]
    &I_{ist} = I_{is(t-1)} + \sum_j P_{ijst} - \sum_c S_{isct} \hspace{2em} \forall i,s,t \hspace{3em} \text{inventory balance constraint} \\[18pt]
    &M_s(1-Z_{ict}) \leq \Gamma_{ict} \leq M_b(1-Z_{ict}) \hspace{8em} \text{outsource amount constraints}               \\
\end{align*}
\end{appendices}

\section*{Acknowledgment}

This research was supported by the Research Foundation-Flanders, grant number 12AZF24N. We are also thankful to Juan Ungredda from our industrial partner for providing the simulator of the practical application.


\end{document}